%% file: main.tex
\documentclass[lettersize,journal]{IEEEtran}
\usepackage{amsmath,amsfonts}
\usepackage{algorithmic}
\usepackage{algorithm}
\usepackage{array}
\usepackage[caption=false,font=normalsize,labelfont=sf,textfont=sf]{subfig}
\usepackage{textcomp}
\usepackage{stfloats}
\usepackage{url}
\usepackage{verbatim}
\usepackage{graphicx}
\usepackage{cite}
\usepackage{booktabs}
\usepackage{enumitem}
\usepackage{arydshln}
\usepackage{multirow}
\usepackage{xcolor}
\usepackage[pagebackref,breaklinks,colorlinks]{hyperref}
\usepackage[capitalize]{cleveref}
\crefname{section}{Sec.}{Secs.}
\Crefname{section}{Section}{Sections}
\Crefname{table}{Table}{Tables}
\crefname{table}{Tab.}{Tabs.}

\begin{document}

\title{Towards Two-Stream Foveation-based \\Active Vision Learning
\thanks{© 2024 IEEE. Personal use of this material is permitted. Permission from IEEE must be obtained for all other uses, in any current or future media, including reprinting/republishing this material for advertising or promotional purposes, creating new collective works, for resale or redistribution to servers or lists, or reuse of any copyrighted component of this work in other works.}%
\thanks{The research was funded in part by CoCoSys, one of seven centers in JUMP~2.0, a Semiconductor Research Corporation (SRC) program.}
}

\author{\IEEEauthorblockN{Timur Ibrayev, Amitangshu Mukherjee, Sai Aparna Aketi, and Kaushik Roy}\\
\IEEEauthorblockA{School of Electrical and Computer Engineering, Purdue University\\
West Lafayette, USA\\
Emails: (tibrayev, mukher44, saketi, kaushik)@purdue.edu
}}

\maketitle

\input{Sections/0-Abstract.tex}

\input{Sections/1-Introduction.tex}
\input{Sections/2-Background.tex}
\input{Sections/3-Methodology.tex}
\input{Sections/4-Results.tex}
\input{Sections/5-Conclusion.tex}

\appendices
\input{Sections/Appendix.tex}

\bibliographystyle{IEEEtran.bst}
\bibliography{main.bib}

\input{Sections/AuthorBios}

\end{document}

%% file: Sections/0-Abstract.tex
\begin{abstract}
Deep neural network (DNN) based machine perception frameworks process the entire input in a one-shot manner to provide answers to both ``\textit{what} object is being observed" and ``\textit{where} it is located". In contrast, the \textit{``two-stream hypothesis''} from neuroscience explains the neural processing in the human visual cortex as an active vision system that utilizes two separate regions of the brain to answer the \textbf{what} and the \textbf{where} questions. In this work, we propose a machine learning framework inspired by the \textit{``two-stream hypothesis''} and explore the potential benefits that it offers. Specifically, the proposed framework models the following mechanisms: 1) ventral (\textit{what}) stream focusing on the input regions perceived by the fovea part of an eye (foveation), 2) dorsal (\textit{where}) stream providing visual guidance, and 3) iterative processing of the two streams to calibrate visual focus and process the sequence of focused image patches. The training of the proposed framework is accomplished by label-based DNN training for the ventral stream model and reinforcement learning for the dorsal stream model. We show that the two-stream foveation-based learning is applicable to the challenging task of weakly-supervised object localization (WSOL), where the training data is limited to the object class or its attributes. The framework is capable of both predicting the properties of an object \textbf{and} successfully localizing it by predicting its bounding box. We also show that, due to the independent nature of the two streams, the dorsal model can be applied on its own to unseen images to localize objects from different datasets.
\end{abstract}

%% file: Sections/1-Introduction.tex
\section{Introduction}
\label{sec:intro}
Current state-of-the-art machine perception systems use deep neural networks (DNNs) to process the entire input image with background \cite{redmon2018yolov3, zong2023detrs} and generate an output without any feedback or the possibility to calibrate itself \cite{he2017mask}.
The training methodologies for such feedforward DNNs have been well explored over the past decade. 
The next milestone is to advance towards solving more complex tasks that humans perform much better than the current machine learning methods, e.g. learning from few examples, visual reasoning, filtering out irrelevant information, etc.
These tasks are challenging because they necessitate designing and incorporating a wide range of human cognitive processes into machine learning systems.
To address this challenge, research interest is gradually shifting towards drawing more inspiration from neuroscience to develop bio-plausible methods that emulate cognitive processes such as learning~\cite{lee2018deep, chakraborty2022action, safa2022stdp}, attention~\cite{mittal2020learning, NEURIPS2021_fc1a3682, tang2023salda}, reasoning~\cite{katz2017novel, lin2017human, zhou2022constructing}, decision-making~\cite{duran2020robot, zhou2022constructing, zhu2022flexible, sanyal2024ev}, and perception~\cite{persson2019semantic, NEURIPS2019_8dd48d6a, nagaraj2023dotie, NeurIPS2020_7866}.
Such a paradigm shift requires machine learning algorithms to be viewed not only in terms of separate tasks but also in their alignment with human cognitive processes as a whole.

One of such interesting ideas on how visual perception works is the ``\textit{two-stream hypothesis}''\cite{goodale1992separate}. 
It suggests that information is processed by two pathways: the dorsal (\textit{where}) and the ventral (\textit{what}) streams. 
The main responsibility of the dorsal stream is to perform visual guidance and identify object locations in space, whereas the ventral stream is responsible for identifying the objects and recognizing fine-grained details.
Although it is still a matter of debate whether two streams are fully independent\cite{sheth2016two}, experimental results have illustrated that each of the streams has distinctive features\cite{norman2002two}.
The dorsal stream processes the input signals received across the entire retina region of the eye, which allows it to perceive the context and object positions in the visual field with respect to the observer.
On the contrary, the ventral stream focuses on the input signals falling onto the fovea region. 
The fovea is a part of the human eye that captures the most amount of details due to the high concentration of cones, a type of photoreceptor with high spatial acuity.
Hence, it allows the ventral stream to process the details of objects and to identify them.

In this paper, we propose \textbf{\textit{a machine learning framework}} inspired by the ``\textit{two-stream hypothesis}'' that uses a combination of DNNs and Reinforcement Learning (RL) schemes. 
In particular, we utilize the following \textbf{mechanisms} from the ``\textit{two-stream hypothesis}'' as the basis for our learning framework.
\textbf{First}, when the input image is received for the first time, the dorsal stream model processes the entire image in low resolution to provide a global image-level context. This mimics processing across the entire retina region and provides the system with a quick initial estimate of the location of the foveal fixation point.
\textbf{Second}, we incorporate the mechanism of foveation of the ventral stream, where a particular glimpse/patch of an image captured by the fovea is perceived in high resolution while the resolution of the remaining image is drastically reduced. In our framework, we enforce foveation with extreme cutoff, where the ventral stream model perceives the input only through a foveated glimpse, and the outlier regions of the input are ignored. This enables the system to focus only on the relevant parts of an object while ignoring the background clutter.
\textbf{Finally}, both the dorsal and ventral stream models process the sequence of foveated glimpses in an iterative manner. At each iteration, the dorsal stream adjusts the fixation point and the size of the foveated glimpse while the ventral stream processes the updated glimpse to extract its features.
In this manner, the system learns to gradually adjust the focus while it actively searches for relevant information from the previously observed parts of a scene.
This is achieved by training the ventral stream as a standard DNN model with image-level information (labels) and using reinforcement learning (RL) to train the dorsal stream on how to adjust the fovea using the output of the ventral stream as a guiding signal (rewards).

\begin{figure*}
    \centering
    \includegraphics[width=0.88\textwidth]{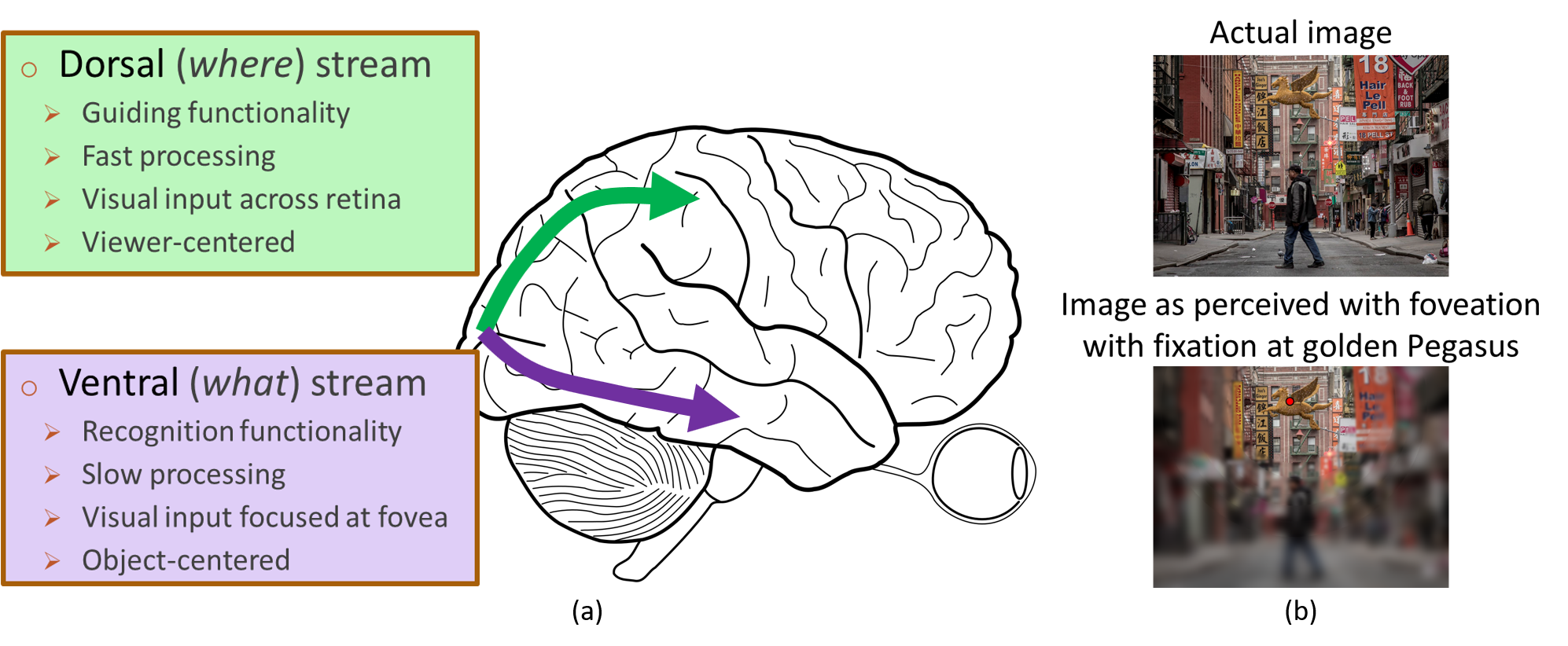}
    \caption{(a) The dorsal (shown in green) and the ventral (shown in purple) stream pathways separate at the primary visual cortex area and stretch to the parietal lobe and the temporal lobe, respectively. 
    (b) Example of foveation in human vision. When looking at the golden Pegasus, the amount of details gradually reduces from the visual fixation point.}
    \label{fig:twostream_and_foveation}
\end{figure*}

The goal of this work is to explore the potential of ``\textit{two-stream hypothesis}'' based learning in guiding the development of new methods for machine perception aligned with the human cognition.
We illustrate the feasibility of the proposed learning framework and its training method for machine perception through the exploratory experiments on weakly-supervised object localization.
\textbf{Weakly-supervised object localization (WSOL)} is the task where the training data only has the object class or its attributes (which are referred to as \textit{image-level labels}), but does not have any information about the ground truth location of the object (which are referred to as \textit{instance-level labels}).
However, during evaluation, the requirement is to predict both the image-level labels \textbf{and} localize the object in the input image by drawing its bounding box (instance-level label).
Two datasets were utilized for this purpose: CelebA\cite{liu2015faceattributes} face attributes dataset, where only object attributes were used for training as image-level labels, and CUB-200-2011 (CUB)\cite{wah2011caltech} bird species dataset, where only class labels were used for training as image-level labels.
We show that the unique training approach of using the standard label-based DNN training for the ventral (what) model and using RL training for the dorsal (where) model makes the proposed two-stream model suitable for solving WSOL tasks.
Specifically, while the mechanism of foveation enables the framework to isolate the object parts from the background clutter, the combined functionality of ventral and dorsal models allows it to capture the entirety of objects from such focused observations.
We also show that, as a direct byproduct of the two streams learning separate functions, the dorsal stream model \textbf{on its own without any re-training} can be used to localize objects in the images of other datasets, namely WIDERFace \cite{yang2016wider} and birds from ImageNet \cite{Deng2009ImageNetAL}.

In summary, this work makes the following contributions:

\begin{enumerate}
    \item We propose a machine learning framework modeling ``\textit{two-stream hypothesis}" with foveation mechanism, which is capable of learning two independent functions -- predicting the object properties and localizing the object location.
    \item We propose a training approach utilizing the combination of label-based DNN training and reinforcement learning.
    \item We verify the feasibility of the proposed framework on the challenging task of weakly-supervised object localization, where the training data is limited only to class labels.
    \item We present the results on CelebA face attributes and CUB-200-2011 bird species datasets, highlighting the capability of the framework to isolate the object parts from the background and to capture the entirety of objects from focused observations.
    \item We also illustrate the generalization capabilities of the framework to unseen images of WIDERFace and the subset of ImageNet datasets.
\end{enumerate}

%% file: Sections/2-Background.tex
\section{Background and related work}
\subsection{Two-stream hypothesis and foveation}
The inception of the idea that the brain utilizes different regions to process the qualities and spatial of objects started in 1969 with the work of Schneider \cite{schneider1969two}. With more supporting evidence and analyses provided by works in \cite{goodale1992separate} and \cite{norman2002two}, the idea morphed into what is now formally known as the ``\textit{two-stream hypothesis}'' (also sometimes referred to as the dual-stream hypothesis) of visual processing.
As shown in Figure~\ref{fig:twostream_and_foveation}, it proposes that there are two pathways in the processing of visual inputs: the dorsal stream and the ventral stream.
The dorsal stream is responsible for visual guidance, meaning that it answers the questions of \textit{where} (according to \cite{schneider1969two, mishkin1983object}) or \textit{how} (according to \cite{goodale1992separate}).
On the other hand, the ventral stream is responsible for visual identification that provides the answer to \textit{what} the object type is. 
While it is still a matter of debate whether or not these two streams are completely independent \cite{sheth2016two}, researchers seem to agree that each of them has its own distinctive features and mechanisms (a comprehensive summary of differences is nicely described in Section 3.4 of \cite{norman2002two}).

One of the mechanisms that we particularly focus on in this work is foveation which mimics the foveal processing detected in the ventral stream, an example of which is shown in \cref{fig:twostream_and_foveation}. 
In contrast to existing object detection/localization frameworks, we explore the possibility of allowing the framework to observe the image only through a series of isolated patches.
Note, by observing the parts of the object, the dorsal stream learns the concepts of these individual pieces and then collates these pieces to recognize the structure of the object in its entirety thus differentiating foreground from background.

The exploration of two-stream processing and foveation in machine learning has been limited.
Authors in \cite{simonyan2014two} proposed a convolutional neural network (CNN) for action recognition from videos, where the information acquired from optical flow data was used to train the temporal network that has similar responsibility as the dorsal stream. 
Work in \cite{peng2016multi} similarly proposed a two-stream object detection architecture, which also combines optical flow information for improved detection. 
In \cite{itti2004automatic}, foveation was shown to be effective for video compression as a mechanism to induce spatial attention via blurring.
More recent work in \cite{deza2020emergent} modeled and analyzed the effects of foveation as a mechanism for efficient and robust encoding in DNN representations.
The proposed framework, to the best of our knowledge, is the first work to explore the capabilities of the ``\textit{two-stream hypothesis}'' and foveation for learning tasks involving self-guided supervision.
The unique training approach allows the framework to learn how to position and adjust the focus on its own.


\subsection{Weakly-supervised object localization}
Object localization/detection plays a very important role in critical systems, such as autonomous vehicles, traffic monitoring, and robotics, to name a few. However, training object detection systems in a supervised manner require massive amounts of costly annotated data. To circumvent this issue, various self-supervised \cite{liu2021unbiased} and weakly-supervised \cite{ren2020instance} approaches have been proposed. In this work, we will show the feasibility of the proposed framework on weakly-supervised object localization (WSOL) where the training data contains the object class or attributes (\textit{image-level labels}) but no bounding box coordinates (\textit{instance-level labels}). WSOL assumes that there is only one object of interest in the entire image.

The works on WSOL can be broadly classified into: CAM~\cite{zhou2016learning} based and Transformer~\cite{vaswani2017attention} based. \\ 
\textbf{\textit{CAM based WSOL:}} Most of the proposed techniques for WSOL utilize and improve the concept of Class Activation Map (CAM) \cite{zhou2016learning} based on classification network to find the object of interest. Since CAM usually highlights the most discriminant parts rather than the entire extent of the object, the works \cite{Mai_2020_CVPR,choe2019attention, choe2020evaluating} focus on improving the CAM by forcing the network to capture the entire object. 
However, all these approaches completely depend on the ability of the classifier and try to localize the objects by learning on the final box obtained as the calibrated CAM outputs. \\
\textbf{\textit{Transformer based WSOL:}} With the recent surge in the use of transformers for visual perception, there have been few attempts at using vision transformers \cite{dosovitskiy2020image} for WSOL and object discovery. The authors of \cite{simeoni2021localizing} (LOST) leverage the activation features of a self-supervised pre-trained vision transformer to find the object in the scene. The key component of the last attention layer in the Transformer is used to compute the similarities between the patches. The patch with the least number of similar patches is chosen as a seed and the patches that are highly correlated to the seed are considered to be part of the same object. LOST localizes the object using patch correlations computed using the features extracted by the pre-trained vision Transformer. TokenCut\cite{wang2022self} is a graph-based technique that uses self-supervised Transformer features to discover an object from an image.

In this paper, we explore an orthogonal approach to WSOL, where the (dorsal) model is forced to learn to actively localize the object. 
Specifically, RL is used to train the dorsal stream model to iteratively calibrate the fixation point and the fovea size. 
The main idea is to adjust the foveated glimpse such that the final glimpse captures \textbf{only} the object and as little background as possible.
The rewards for RL are assigned based on how well outputs from the currently observed foveated glimpse (obtained from the ventral stream model) match the expected image-level targets.
The \textit{main distinction of our framework} is that in order to learn the correct sequence of actions to capture the object, it has to learn the underlying structure of an object.
For example, if the current glimpse captures the eyes in the face recognition task, the dorsal stream model predicts expanding the glimpse down to capture the nose and mouth parts of the face.
Following that, the ventral stream model will use the updated glimpse, which now includes eyes, nose, and mouth, to identify the face attributes.
Moreover, the proposed training also allows the dorsal stream model to be used independently on its own after the training is completed since two models learn independent functions.
This is in contrast to previously mentioned WSOL approaches, which re-purpose the classifier, and are trained to predict the coordinates of the object in a single step.
Such detectors rely on the fact that the object of interest is always observed in its entirety. 
Hence, they are not capable of \textbf{actively} making decisions based on the underlying structure of the object.
For example, if these detectors observe the left part of the face, they will just predict that it is indeed a (whole) face and will not give the information to search to the right to potentially observe the entire face.

Note that even though we also make use of the Grad-CAM \cite{selvaraju2017grad} technique, it is only used during training of the dorsal stream model to find the initial point of interest in the input scene.
Also, the idea of using reinforcement learning for active localization has been explored in the works of \cite{caicedo2015active, mnih2014recurrent, ba2014multiple}. 
However, none of the previous works focus on localization in the WSOL setting without bounding boxes.

%% file: Sections/3-Methodology.tex
\begin{figure*}[ht]
    \centering
    \includegraphics[width=0.87\textwidth]{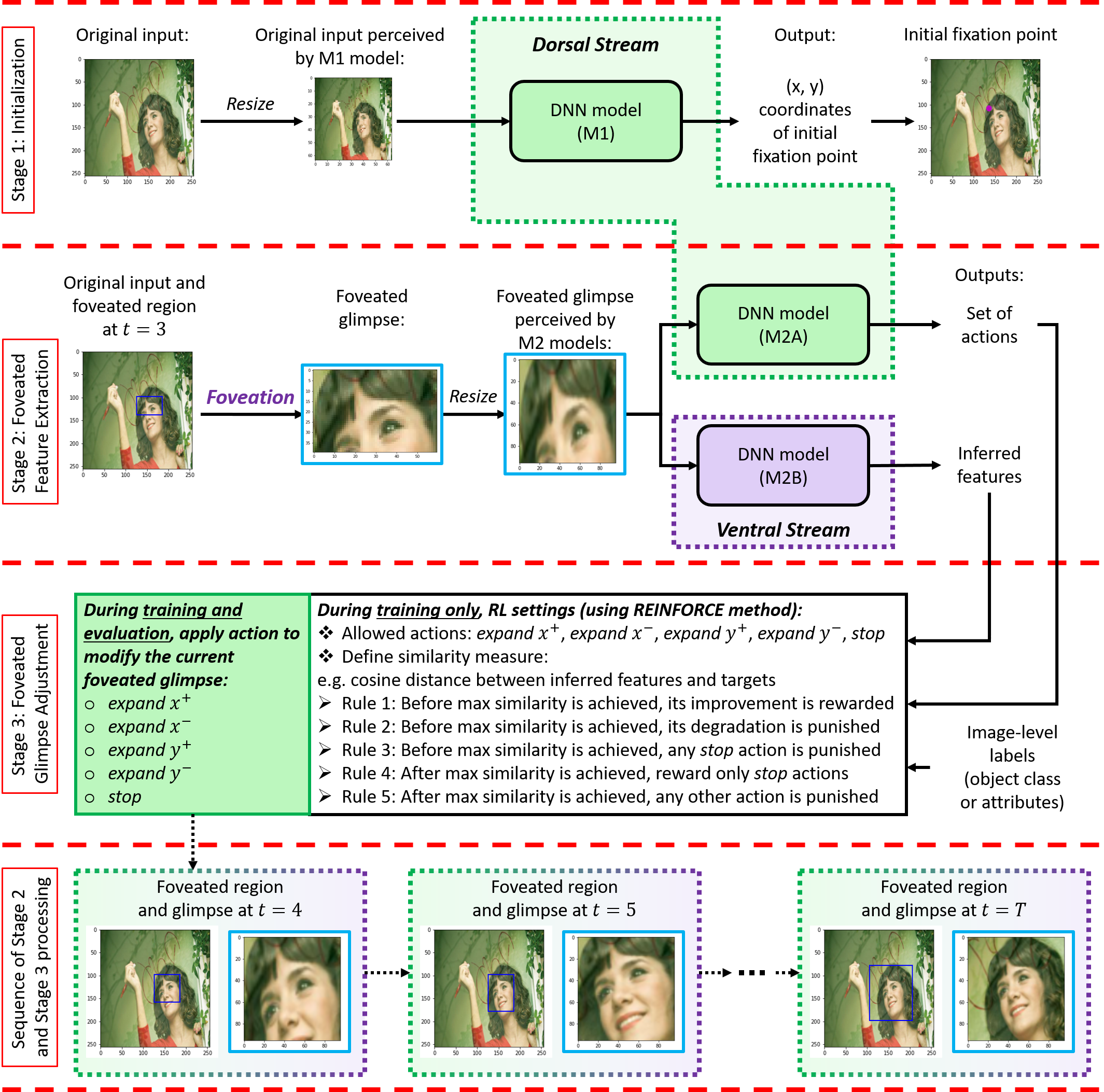}
    \caption{Overview of the proposed framework. \textbf{Stage 1} (initialization) takes place when the input is observed for the first time and uses the dorsal M1 model to predict the initial fixation point. \textbf{Stage 2} (foveated feature extraction) processes the foveated glimpses using the dorsal M2A and the ventral M2B models to predict the set of actions to adjust the foveation, and extract features identifying the object of interest, respectively. The example shows foveated glimpse extracted at the 3rd iteration. \textbf{Stage 3} (foveated glimpse adjustment) adjusts the foveated region (i.e. the blue rectangle shown on the input image) resulting in a new foveated glimpse. \textbf{Stages 2 and 3} are then iteratively implemented to process and extract the sequence of foveated glimpses with the goal of localizing and identifying the object in the image.
    \vspace{-1.5mm}}
    \label{fig:main_framework}
\end{figure*}

\section{Methodology}
\Cref{fig:main_framework} shows the overview of our framework consisting of three models M1, M2A, and M2B. 
By training these models, the entire process can be divided into three main stages: 1) initialization (by M1), 2) foveated feature extraction (by M2B), and 3) foveated glimpse adjustment (by M2A).
While the initialization stage occurs only once for a given input, stages 2 and 3 are performed iteratively on a sequence of resulting foveated glimpses.
Next, we describe the details of each stage with their realization and training.


\subsection{Stage 1: initialization}
When an input is observed for the first time, the initial function performed by our framework is that of the dorsal stream. 
Specifically, it mimics the dorsal stream to process the visual signal falling onto the entire retina area to provide an initial visual guidance based on the global context. 
This is modeled by a DNN (M1 shown in ~\cref{fig:main_framework}) that processes an image in low resolution and outputs the coordinates $(x,~y)$ of the initial predicted fixation point.
There are two reasons to process the input image in low resolution. 
First, it intentionally reduces the amount of details that the input image contains. Note, the main responsibility of this stage is to find an \textit{approximate} estimate of the location of the object of interest while not paying strong enough attention to identify the object. Second, it allows for a compact DNN (i.e. neural networks with fewer layers, each having few channels) for the realization of M1. 
This results in faster processing by the model (in terms of compute operations) while still taking the entire image-level global context into account.

The M1 model is trained in a supervised manner with a regression (mean squared error) loss. If there are bounding boxes provided by the target task, M1 can be trained to predict their centers. 
However, in the case of a WSOL task, the information about the location of the object in the scene is not available.
Hence, \textit{for the purpose of training M1 model}, we rely on the Grad-CAM technique \cite{selvaraju2017grad}, a technique of visualizing the coarse regions of an input that are ``important'' for the predictions by \textbf{any trained model}. 
It produces the heatmap indicating the parts of the input which had the highest effect on the outputs of the model being analyzed.
The training of the M1 model using Grad-CAM is implemented in three steps.
\textbf{In the first step}, we train a separate model on image-level labels of the target dataset. 
The resulting model is referred to as the \textit{pre-trained model}. 
The pre-trained model and M1 model do not have to have the same network architecture. 
Contrary to the simple DNN for M1, we use a complex DNN for the pre-trained model that captures all the image-level details of the target dataset.
\textbf{In the second step}, the Grad-CAM technique is applied to the pre-trained model to get the ``attention" heatmap for every image in the training set, indicating which part of an image was the most significant in producing the corresponding outputs. 
\textbf{In the final step}, the values of each output heatmap are then treated as the probabilities of an object location in the images. 
The (pixel) location with the highest probability value for each image is then used as \textbf{the target label to train M1 network}. 
As a result, the M1 model is trained to output the coordinates $(x,~y)$, which are then used as initial fixation points. 
It is not expected for the M1 model to predict the exact target locations extracted from the ``attention" heatmaps. 
Contrary, since its main function is to provide an approximate object location, any prediction falling within the proximity of this location in the input space is considered useful.


\subsection{Stage 2: foveated feature extraction}
\label{sec:foveation}
After the visual guidance in the form of the initial fixation point is obtained from the dorsal stream (Stage 1), our framework implements the foveation mechanism believed to be the part of ventral stream processing.
Parts of the input visual signals falling onto the fovea of an eye are processed with more details. The amount of details processed gradually reduces across the visual field from the fixation point of the eye. 
The way our framework models foveation is by extracting (cutting out) the region of the input that needs to be processed in high detail and completely disregarding the rest of the input.
When the input is processed for the first time after Stage 1 (initialization), foveation is applied by extracting the region of some fixed size centered at the coordinates $(x,~y)$ of the initial fixation point predicted by the M1 model. 
The resulting patch is the first \textbf{\textit{foveated glimpse}}.
All of the foveated glimpses are then resized to a predetermined fixed size regardless of their actual sizes and aspect ratios. 
This is performed in order to simultaneously process a batch of foveated glimpses, which later will be of different shapes as a result of iterative processing during training (explained in the next subsection).

The resized foveated glimpse is processed by a DNN (M2B shown in \cref{fig:main_framework}) that models the functionality of the ventral stream. 
Similar to the ventral stream, M2B model extracts and outputs features corresponding to the input foveated glimpse that can be used to identify the object or its parts.
To achieve this, we initialize M2B model using the weights of the pre-trained model used in the training stage of M1 model and is trained only on image-level labels.
Although it is potentially possible to train the M2B model from scratch or by fine-tuning it on the foveated glimpses, we only consider the M2B model with weights loaded and frozen from the pre-trained network in all our experiments.


\subsection{Stage 3: foveated glimpse adjustment}
\label{sec:rl_training}
As the ventral stream model extracts features to identify what object is being observed, the same foveated glimpse is processed by another DNN model (M2A shown in \cref{fig:main_framework}). 
M2A model mimics the adjustment of the focus size, by iteratively adjusting the dimensions of the foveation region.
The M2A model achieves this by predicting one of the \textit{actions} which is then performed to adjust the foveated glimpse.
The set of allowed actions includes expanding the foveation along one of the four directions or keeping it unchanged.
In other words, the M2A model guides the fixation point and adjusts the spatial attention.
Since the functionality of both M1 and M2A models is to provide the visual guidance, both of them collectively model the dorsal stream.
However, while M1 model predicts the initial fixation point based on the global context, M2A model adjusts the focus size based on the latest local information.

\begin{figure*}
    \centering
    \includegraphics[width=0.96\textwidth]{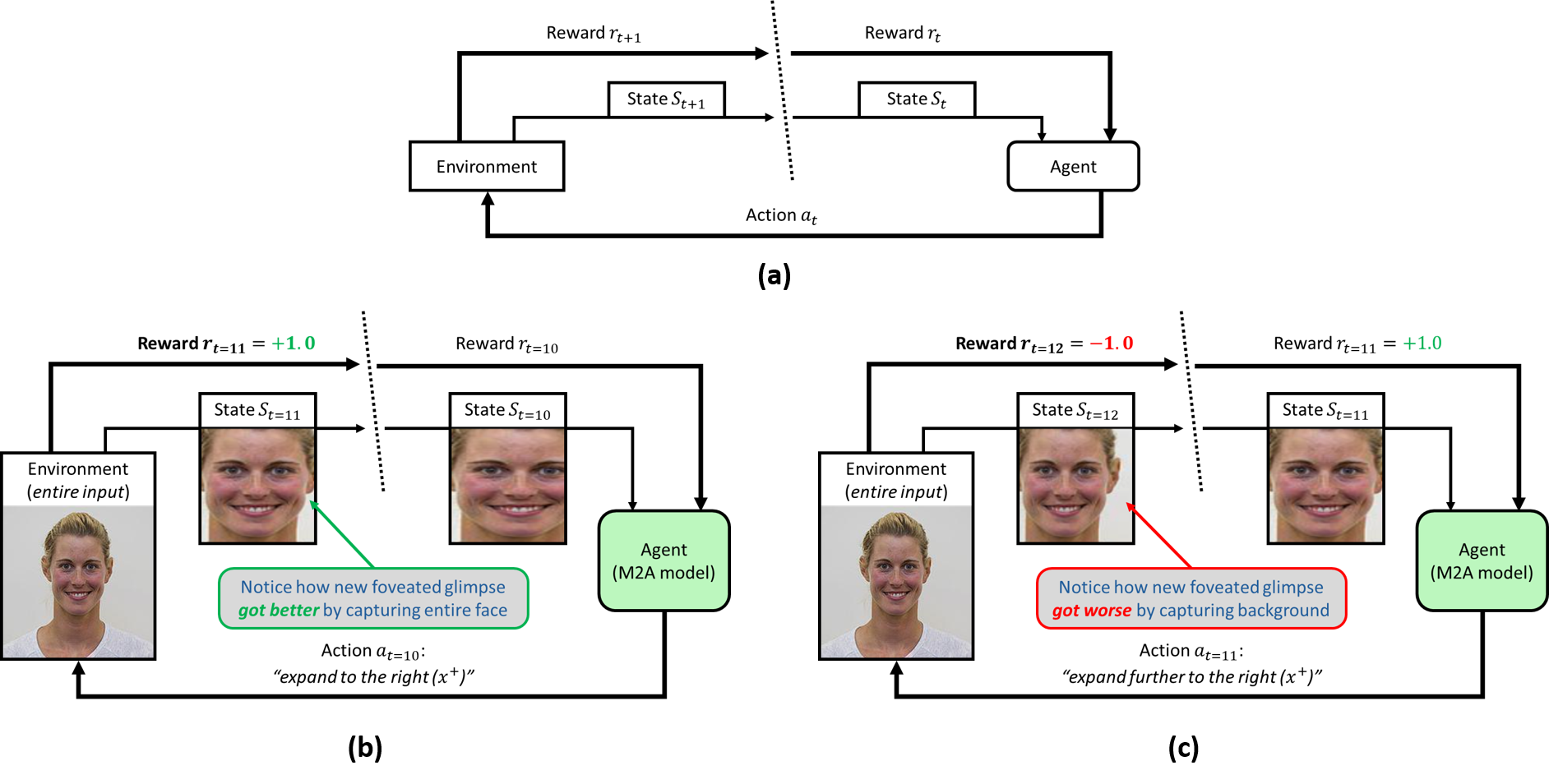}
    \caption{Overview of how the M2A model training is formulated using reinforcement learning. (a) Standard scheme of reinforcement learning, illustrating how the environment provides state and reward information based on the action that an agent performs on the environment. (b) Example of how M2A model (i.e. the agent) performs the action ($a_{t=10})$) on the environment, which result in \textbf{a positive improvement} in the form of improved localization from the state $S_{t=10}$ to the state $S_{t=11}$. (c) Example of how M2A model (i.e. the agent) performs the action ($a_{t=11})$) on the environment, which result in \textbf{a negative/undesired change} in the form of unnecessarily expanding the foveated glimpse from the state $S_{t=11}$ to the state $S_{t=12}$.}
    \label{fig:rl_training}
\end{figure*}

Once the M2B model determines the features observed from the current foveated glimpse and the M2A model determines the optimal action to enrich observed features, a bottom-up processing is performed in the form of adjusting the foveation based on the predicted action. (Although we refer to the adjustment of the foveation as the bottom-up process, please note that it does not involve any additional computation. Rather, it involves a direct execution of actions predicted by the M2A model.)
The sequence of processing new foveated glimpses and adjusting the foveation to expand the visual input received by the M2 models is then repeated for a predetermined \textit{fixed} number of iterations.

In the WSOL setting, at each iteration the M2B model determines the features of the object present in the image and the M2A model predicts the adjustments to the foveation. 
The localization task is then for the final foveated glimpse to capture the entirety of the object.
The final foveated glimpse should entirely capture the object. 
Its dimensions and position in the input image (i.e. the foveated region) are then used to evaluate the performance of the framework on the WSOL task.


Reinforcement learning (RL) is used to train the M2A model to predict the correct sequence of foveated glimpses through the correct sequence of actions to adjust the foveation.
The advantage of using RL is that it allows not only for the foveation to be an active decision making process, but also for this process to be effectively learnt by the M2A model itself through trial and error.
\Cref{fig:rl_training}(a) illustrates the general overview of the reinforcement learning setting with its components. The general scenario is for an agent to act on the environment through some action $a_t$ based on the previous observation of some state $S_t$ originating from the environment. Based on the environment and the performed action $a_t$, the agent observes the state change from $S_{t}$ to $S_{t+1}$ and receives the reward/punishment $r_{t+1}$ for its action.
We use REINFORCE reinforcement learning algorithm~\cite{williams1992simple}. 
This is a Monte-Carlo method for policy optimization, where a policy refers to the decision making of the agent.
The algorithm trains the agent by treating its output predictions as probabilities of each action.
The training process of the algorithm is to alternate between allowing the agent to perform a sequence of such stochastic actions based on its current policy and optimizing the policy based on how close the sequence of actions matched the desired behavior.
As a result, the optimization involves increasing the probability of actions which were desirable and reducing the probability of actions which made the agent deviate from the expected behavior.

In our framework, the M2A model acts as the agent. 
It means that the agent policy is parameterized by the M2A model weights and the model outputs represent the probability of each of the allowed actions.
The set of actions is limited to $5$: 
\begin{enumerate}[nosep]
    \item expand the left border of the glimpse ($expand~x^-$)
    \item expand the right border of the glimpse ($expand~x^+$)
    \item expand the top border of the glimpse ($expand~y^-$)
    \item expand the bottom border of the glimpse ($expand~y^+$)
    \item keep the glimpse unchanged ($stop$)
\end{enumerate}
The entire input image serves as the environment and foveated glimpses perceived by M2A and M2B models at each iteration act as states.
The first state $S_{t=1}$ for every input is the first foveated glimpse, which is the region of some predetermined fixed size centered at the $(x,~y)$ coordinated predicted by the M1 model as the initial fixation point (as described in \Cref{sec:foveation}).
Then, the consecutive states are determined by the next foveated glimpses, which are obtained as a result of applying one of the actions predicted by the M2A model.
Formally, the state $S_{t+1}$ (the new foveated glimpse) is obtained based on the action $a_{t}$ that had the highest output probability produced by the M2A model as a result of processing the state $S_{t}$ (the previous foveated glimpse).

The final component for RL training of the M2A model is the reward function, which instructs the algorithm how to assess each predicted action.
In the current version of the framework verified on the WSOL task, only the ventral stream M2B model receives a direct supervision during training.
Hence, the reward function plays the vital role in training the dorsal M2A model to perform the localization of the object of interest from the observed glimpses.
Based on this, the main assumption that is used in designing the reward function is that the ground truth location of the object contains the most optimal information about the object properties.
In other words, we assume that when the entire object is observed separately from the background, the ventral stream model, M2B, produces outputs that are the most similar to the image-level labels.
Similarity measures can either be the output confidence in the target class or the cosine distance between the predicted attributes and the ground truth attributes of the object. 
The similarity measure can be used to assign appropriate rewards based on the following five rules:
\begin{enumerate}[nosep, label={(\arabic*)}]
    \item Before max similarity, an improvement in similarity is rewarded
    \item Before max similarity, a degradation in similarity is punished
    \item Before max similarity, any \textit{stop} action is punished
    \item After max similarity, reward only \textit{stop} actions
    \item After max similarity, any other action is punished
\end{enumerate}

\Cref{fig:rl_training} shows examples of training the M2A model training using REINFORCE algorithm and the described training components. 
\Cref{fig:rl_training}(b) illustrates the positive reinforcement of the M2A model. 
The action $a_{t=10}$ improves the foveated glimpse by capturing the entirety of the object (in this case a human face). 
Since the localization (and hence, the similarity with the image-level label) improves from the state $S_{t=10}$ to the state $S_{t=11}$, the positive reward $r_{t=11}$ is provided to the M2A model for the action $a_{t=10}$.
 ~\Cref{fig:rl_training}(c) illustrates the negative reinforcement of the M2A model. 
The action $a_{t=11}$ excessively expands the foveated glimpse to capture a part of the background. 
Note, how state $S_{t=12}$ includes a part of the input image which does not describe any of the facial attributes. 
Consequently, the negative reward $r_{t=12}$ is provided to the M2A model for the action $a_{t=11}$, which reduces the probability of repeating this action by the REINFORCE algorithm.

Following this training method, the M2A model learns to adjust the fovea size by expanding the foveated glimpse while the similarity is increasing and to keep it unchanged when the similarity saturates or starts to reduce. 
The above rules are general for M2A training, but the precise reward values have to be fine-tuned based on the target dataset.
During inference, the framework iteratively applies the learned actions, \textbf{without the need to evaluate the similarity of the M2B outputs and the image-level labels}. The action that corresponds to the highest output value produced by M2A is chosen to alter the foveated glimpse.

%% file: Sections/4-Results.tex
\begin{figure*}[ht!]
    \centering
    \includegraphics[height=0.47\textwidth]{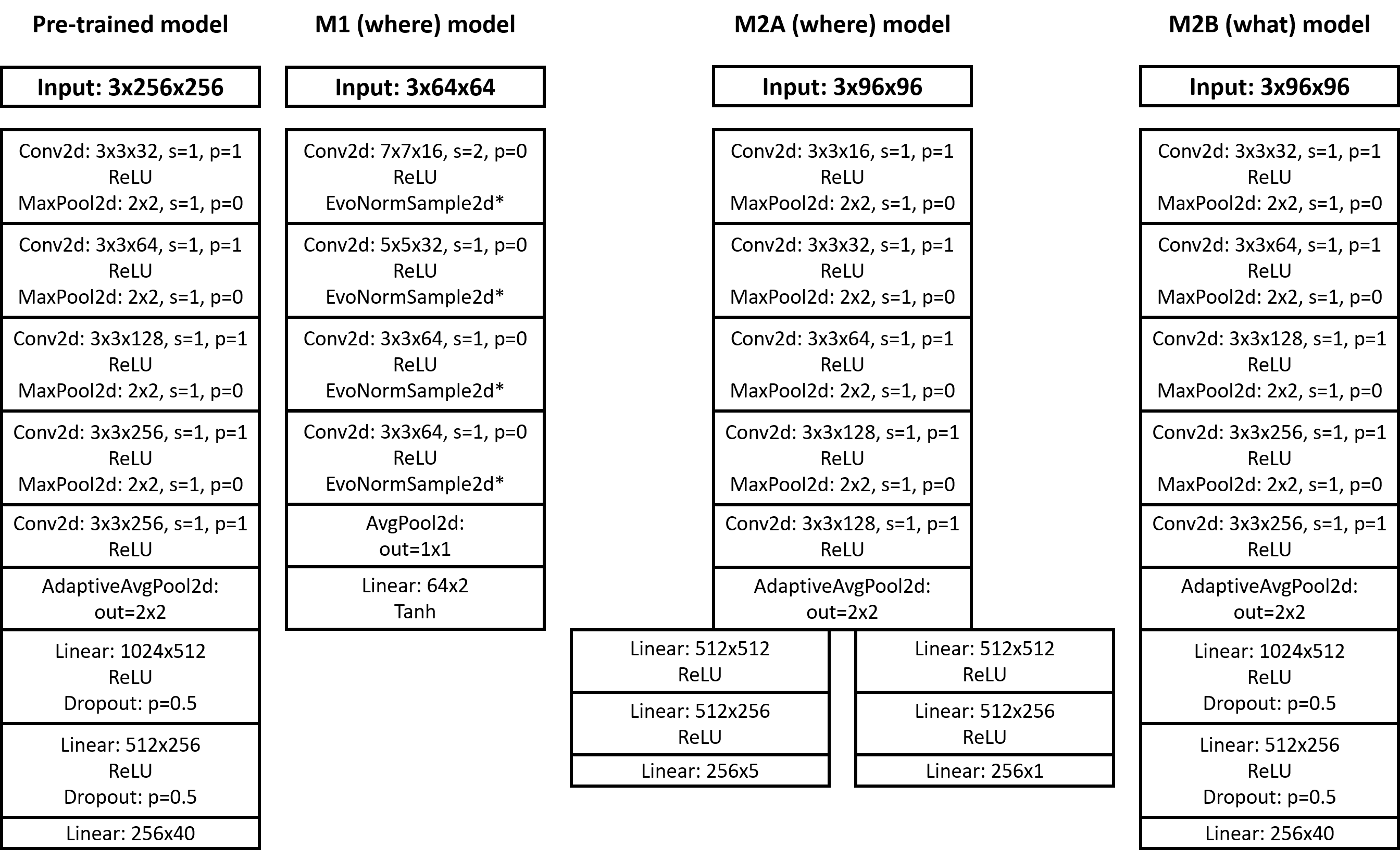}
    \caption{Neural network architectures used to realize different models needed by the proposed framework for experiments on CelebA dataset.}
    \label{fig:celeba_models}
\end{figure*}

\section{Experiments \& Results}
\label{sec:results}
\subsection{Overall Setup}
\label{sec:exp_setup}
The training of the proposed framework can be accomplished by only partial supervision. 
Hence, it felt natural to explore its applicability to weakly-supervised object localization task.
Specifically, we investigate training the framework on two datasets: CelebA\cite{liu2015faceattributes} and CUB-200-2011 (CUB/birds)\cite{wah2011caltech} datasets.
\Cref{sec:res_celeba,sec:res_cub} describe the experiments for each dataset.
Moreover, in \cref{sec:res_generalizability} we show that the dorsal stream model, M2A, indeed learns to independently localize the objects akin ``\textit{two-stream hypothesis}". 
We demonstrate that after training on CelebA and CUB datasets, dorsal M2A model alone can be successfully used to localize faces of WIDERFaces\cite{yang2016wider} and birds of ImageNet\cite{Deng2009ImageNetAL} datasets, respectively.

We present the results on test splits of each dataset achieved with the parameters tuned based on validation splits of the corresponding datasets.
For each dataset, we describe experimental details before discussing the quantitative and qualitative results.
Specifically, each of the following sections presents the neural network models, the RL reward functions, and the training hyperparameters used for the corresponding dataset. 
Other qualitative results are presented in the appendix.
Please note, however, that the methodology and the rules of training RL described in \cref{sec:rl_training} remain the same.
The entire framework was implemented using PyTorch framework \cite{NEURIPS2019_9015}. 
The models were trained and evaluated using NVIDIA GeForce GTX 1080 Ti GPU card.


\subsection{CelebA dataset}
\label{sec:res_celeba}
\subsubsection{\textbf{``In-the-wild" uncropped images}}
The feasibility of the proposed framework with foveation for active object localization task was initially verified by training and evaluating on the CelebFaces Attributes (CelebA) dataset \cite{liu2015faceattributes}.
CelebA consists of $202599$ face images each annotated with 40 binary attributes. 
The original CelebA dataset contains two different versions: the ``in-the-wild'' images and the ``aligned and cropped'' images.
For the tasks like image generation, usually the ``aligned and cropped'' images are used. These are essentially the same images as the ``in-the-wild'' images, which were aligned according to the two eye locations and cropped to the same size of $218\times178$ pixels.
As a result, ``aligned and cropped'' images can be considered as already localized on the face region.
One of the goals of our experiments was to analyze how foveation helps to distinguish between background clutter and the object of interest. Thus, we only used the ``in-the-wild'' images in all experiments. The bounding boxes that were used to align and crop the ``in-the-wild'' images are then used to evaluate the localization performance.

\subsubsection{\textbf{Neural network model architectures}}
\Cref{fig:celeba_models} illustrates neural network architectures used to realize different models in the proposed framework for CelebA experiments. 
The framework performs inference using the three models (M1, M2A, and M2B).
Note, we also describe the architecture of the pre-trained model, which is used only during training for two purposes: (1) it is used to get targets to train on the M1 model (using Grad-CAM technique, refer to Stage 1 description), and (2) it is then used to initialize the weights of M2B ventral (what) model.

In addition, it should be highlighted that models have different input dimensions. While the pre-trained model is trained on large input size ($256\times256$ RGB images), the models of the framework receive inputs of smaller dimensions.
As described in Stage 1 description, the M1 model processes inputs with a reduced resolution of $64\times64$ pixels.
As described in Stage 2 description, M2A, and M2B models process each foveated glimpse resized to the same fixed size of $96\times96$ pixels.
Moreover, the M2A model has two RL heads.
The head with 5 output nodes is used for predicting one of the allowed actions.
The head with the single output node is used for predicting a baseline value, which is used in the policy optimization methods to stabilize the training. Note, the baseline head can be discarded during the test.

\begin{figure*}
    \centering
    \includegraphics[width=0.9\textwidth]{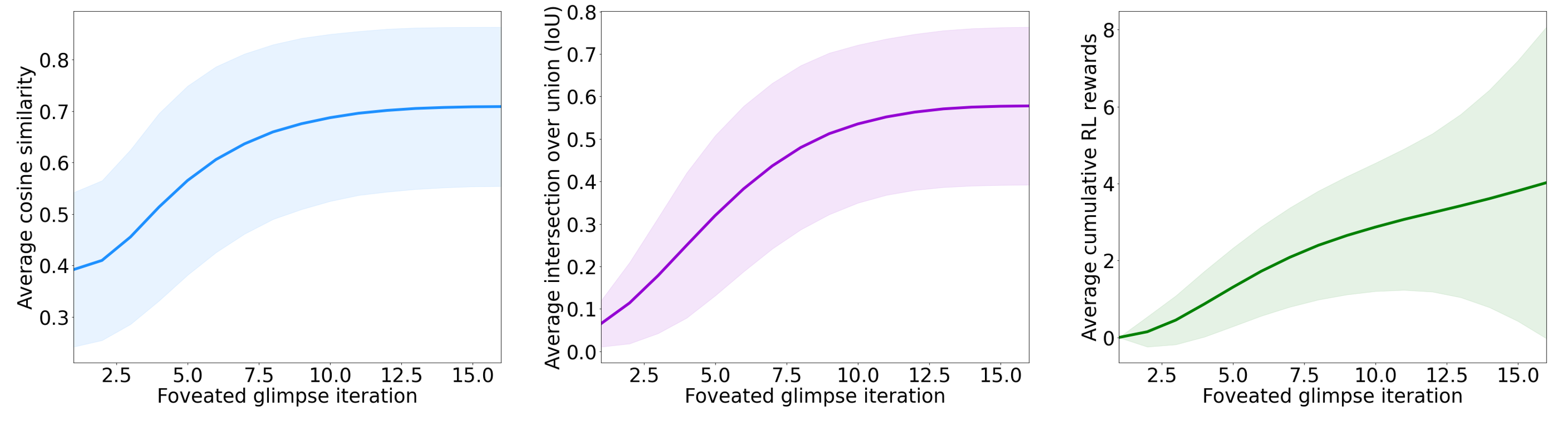}
    \caption{Statistics over the foveated glimpse iterations: (top left) average cosine similarity, (top right) average IoU, (bottom) average cumulative reward for RL averaged over test split of CelebA dataset.}
    \label{fig:stats_over_iterations}
    \vspace{-0.5cm}
\end{figure*}

\subsubsection{\textbf{Reward assignment for RL}}
The training of the M2A model using reinforcement learning with the methodology explained in \cref{sec:rl_training} depends on (1) the selected similarity measure and (2) the rules determining how to assign rewards based on the selected similarity.
CelebA dataset has attributes as the image-level labels, which is a $40$ dimensional vector.
Hence, for the experiments on the CelebA dataset, we used cosine distance as the similarity measure.
In particular, cosine distance is measured between $40$ dimensional target attribute vectors and the output predictions of M2B model of the same dimensions.

During training, we would allow the foveated glimpses to expand according to the actions predicted by the M2A model and process each of the resulting glimpses by the M2B model.
As a result, over the full course of glimpse iterations (also known as one RL trajectory), we would collect the similarity measures corresponding to each glimpse and resulting from each action.
Then, each action is assigned the reward based on the similarity between M2B model predictions and the target labels. 
For experiments on the CelebA dataset, the reward assignment exactly follows the five rules described in \cref{sec:rl_training} and as depicted in \cref{fig:main_framework}:
\begin{enumerate}[nosep, label={(\arabic*)}]
    \item Before max similarity is achieved, an improvement in similarity is rewarded ($reward=1.0$)
    \item Before max similarity is achieved, a degradation in similarity is punished ($reward=-0.25$)
    \item Before max similarity is achieved, any \textit{stop} action is punished ($reward=-1.0$)
    \item After max similarity is achieved, reward only \textit{stop} actions ($reward=1.0$)
    \item After max similarity is achieved, any other action is punished ($reward=-0.25$)
\end{enumerate}
The values for the rewards were determined based on the performance on the validation split of the dataset. However, they were not exhaustively verified for optimality, meaning that there might be a set of rewards that yields better localization performance.

\Cref{fig:stats_over_iterations} illustrates the change in key statistics during the iterative processing of the foveated glimpses averaged over the test split of CelebA. 
As expected, both the average cosine similarity as well as the average IoU monotonically increase as the iterations progress, indicating the fact that the predicted adjustments localize the face regions to correctly identify face attributes. 
A large spread of average cumulative RL rewards towards the end of the iterations highlights that some of the actions become redundant/unnecessary according to our defined rules.
A closer look at the defined rules might reveal the cause of this behavior. Specifically, the dorsal M2A model predicts additional moves outside the face region in order to make sure there is no additional information relating to the face attributes.

\input{Tables/Hyperparams_CelebA}

\subsubsection{\textbf{Hyperparameters}}
\Cref{tab:celeba_hyperparams} shows the hyperparameters used for CelebA experiments.
During all stages model parameters are optimized using Adam optimizer~\cite{kingma2014adam}. 
The important hyperparameters pertaining to our proposed framework are: $num\_glimpses$, $fovea\_control\_neurons$, $glimpse\_size\_init$, $glimpse\_size\_fixed$, and $glimpse\_size\_step$. 
$num\_glimpses$ is the total number of times stage 2 and stage 3 need to be performed iteratively. It means the framework has to process this number of foveated glimpses and correctly predict the same amount of actions to adjust them. $fovea\_control\_neurons$ then defines the number of actions that the framework is allowed to perform. Importantly, these include $stop$ action for the cases when foveated glimpse already localized the face, but there are still iterations required to be performed.

$glimpse\_size\_init$ is the size of the very first foveated glimpse, which is centered around the initial fixation point predicted by the M1 dorsal model.
$glimpse\_size\_step$ determines the step size in $(x,~y)$ coordinates by which foveated glimpses can be adjusted during stage 3 based on the predicted action. For example, if the predicted action is $expand~x^-$, then the left border of the glimpse is expanded to the left by $glimpse\_size\_step$ amount of pixels. 
It is important to remind that the current implementation of the framework only supports expand actions, and the set of 5 allowed actions independently takes care of each glimpse border (refer to the set of actions described in stage 3 description).
Every foveated glimpse is resized to a fixed size determined by $glimpse\_size\_fixed$ before being processed by M2A and M2B models. This includes both the glimpses which are smaller than this size, which means glimpses are stretched, and the glimpses that are larger than this size, which means glimpses are shrunk.

\subsubsection{\textbf{Quantitative and qualitative results}}
\Cref{tab:celeba} shows the performance of each of the individual components and the overall system on object identification and its localization using 4 different accuracy measures. 
First, the \textit{attribute accuracy} measures the percentage of correctly predicted attributes. Because the CelebA dataset does not have class labels, the accuracy of predicting attributes serves as the proxy of the capability of the framework to identify the object's qualities. 
Next, the \textit{localization accuracy} is reported as two values: (a) hit/miss accuracy and (b) ground truth-known localization (GT Loc).
Hit/miss accuracy is the percentage of the total number of samples for which the M1 dorsal model's predicted initial fixation point falls within the bounding box. 
Ground truth-known localization (GT Loc) reports the percentage of the predicted bounding boxes that have the intersection over union (IoU) with the ground truth bounding box of more than $0.5$. 
Finally, the \textit{attribute localization accuracy} is the combination of the \textit{attribute accuracy} and \textit{GT Loc}: the percentage of correctly identified attributes only when the face is correctly localized.
The results of the performance of the entire framework are reported for 8 training runs using different initializations: 3 are trained on both train and validation splits and 5 are trained only on train split.
The reported values are mean and standard deviation.
The proposed framework was capable of localizing the face region in $72\%$ of cases. 
In turn, this resulted in $64\%$ of attributes being both correctly identified and localized.

\begin{figure}
    \centering
    \includegraphics[width=0.95\columnwidth]{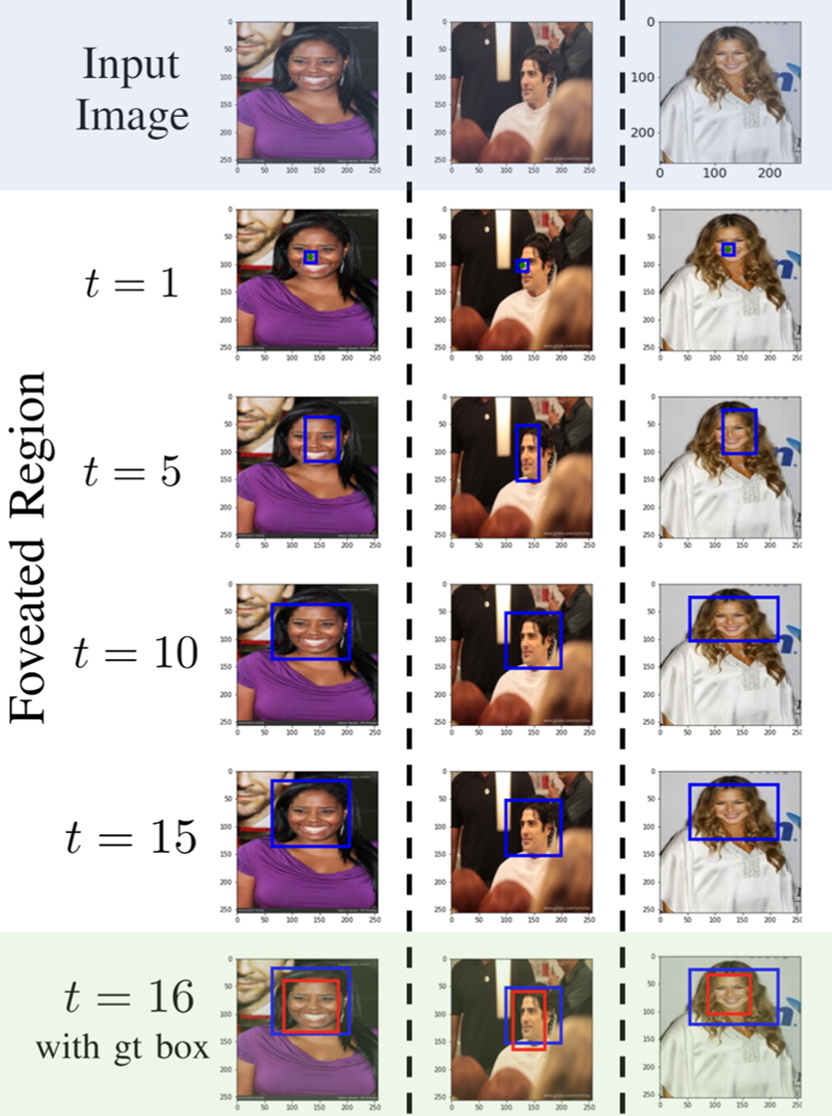}
    \caption{Randomly chosen examples from the test split of CelebA. 
    Blue rectangles show the foveated glimpses. It means that the framework only observes the parts of the input within these blue regions \textbf{and} on its own and predicts how to calibrate them to localize the object (face). Red rectangles show the ground truth bounding boxes provided by the dataset, but which are not used by the framework.\vspace{-12pt}}
    \label{fig:test_samples}
\end{figure}

\Cref{fig:test_samples} shows a few randomly chosen examples from the test split of CelebA and the corresponding sequence of the foveated regions that the framework extracts and perceives to achieve the final localization. More sample images from CelebA dataset are provided in the appendix.

\input{Tables/Results_CelebA}

\subsection{CUB-200-2011 dataset}
\label{sec:res_cub}

\subsubsection{\textbf{Neural network model architectures}}\label{sec:res_cub_nnarchs}
For the experiments on CUB-200-2011 (birds) dataset, M1 model remained the same as shown in \cref{fig:celeba_models}, whereas the network architectures of the pre-trained, M2A, and M2B models were changed.
Due to the dataset complexity, the above mentioned components of the framework were realized as ResNet50\cite{comptable_resnet50} deep neural network.
When training on the CUB-200-2011 dataset, the weights of the convolutional layers of the pre-trained model were initialized to those trained on the ImageNet dataset.
They were then fine-tuned and the classifier was trained on the image-level labels.

\subsubsection{\textbf{Reward assignment for RL}}
The general methodology that we used for RL training remains the same as described in \cref{sec:rl_training}.
However, based on the differences between CelebA and CUB datasets, some adjustments were made to both (1) the selected similarity measure and (2) the rules determining how to assign rewards.

CUB dataset has both class labels and attributes as the image-level labels. However, in this work, we attempted to focus only on using class labels for the CUB dataset. There are two reasons for that: (1) we wanted to see how the proposed framework will behave only on class label information and (2) to have as close a comparison as possible with other WSOL algorithms.
Hence, during the training of the proposed framework, we used the confidence in the correct class as the similarity measure.
In particular, the output predictions from the M2B model were passed through the softmax layer and then the value that corresponds to the target class was taken as the similarity of the current foveated glimpse.
This ensured that only the information about the correct target class is used to guide the actions of M2A.

The change in the selected similarity measure necessitated changes in the reward assignment.
One of the main changes is in determining the maximum similarity, which essentially separated rules (1-3) and rules (4-5) in \ref{sec:rl_training}.
CelebA attributes allowed quantized changes in the similarity measure. It means that there needs to be a substantial change in the foveated glimpse in terms of inferred information, in order to flip one output in the $40$ dimensional output prediction vector of the M2B model. 
In contrast, any change in the foveated glimpse causes the change in the output confidences produced by the M2B model.
Hence, the maximum similarity was changed to be the first highest similarity that is greater than the current one \textbf{by some predetermined percentage}. 
In other words, if the previous (in case of the CelebA dataset) assignment of the max similarity ($maxS$) can be described by \cref{eq:celeba_max}, then now (in case of the CUB dataset) the assignment changed to \cref{eq:cub_max}. 
The value of the predetermined percentage (i.e. $sim\_change\_th$) was based on the observations made on the validation split. 
Adding this threshold in determining the maximum similarity essentially allowed to discard small variations in the similarity, especially when the similarity reached larger values.

\begin{equation}
\label{eq:celeba_max}
    maxS \leftarrow S_t~\texttt{if}~S_t \geq maxS
\end{equation}

\begin{equation}
\label{eq:cub_max}
    maxS \leftarrow S_t~\texttt{if}~\frac{(S_t - maxS)}{maxS} \geq \textit{sim\_change\_th}
\end{equation}

The rules were also augmented with additional conditions. 
Specifically, we added a penalty to the RL actions if over the entire course of actions (RL trajectory), the maximum similarity was \textbf{less than some predetermined value} ($sim\_min\_satisfactory$).
This was necessary to prevent the M2B model to receive rewards even if any parts of the object are not found.
In this situation, all of the actions received (maximum) negative penalty, discouraging M2B model to produce similar useless trajectories.

\input{Tables/Hyperparams_CUB}

\subsubsection{\textbf{Hyperparameters}}
CUB dataset by default has only train and test splits. Hence, we divided the train split into train and validation splits, by randomly selecting $3$ images per class from the train split.
Then, the hyperparameters were selected based on the performance of the framework on the validation split.
After hyperparameters were determined, the training was completed on the entire training split and the model performance evaluated only on the test split.

\Cref{tab:cub_hyperparams} shows the hyperparameters selected for CUB experiments.
While SGD optimizer was used for fine-tuning the pre-trained model and training the M2A model, Adam optimizer was used for training the M1 model.
All of the hyperparameters have the same functionality as described for CelebA experiments.
Two new hyperparameters added due to the change in reward assignment for the CUB dataset are:
$sim\_change\_th$ and $sim\_min\_satisfactory$.
$sim\_change\_th$ determines how much the similarity of any iteration should be larger than the current highest similarity in order for it to become the new highest similarity. This value is treated as the percentage of the current highest similarity. Note, a value of $0.1$ represents that the similarity should be larger than the current highest similarity by $10\%$ in order for it to become the new highest value.
$sim\_min\_satisfactory$ determines the minimum highest similarity value that the foveated glimpses should have in order for M2A actions to be considered useful. Otherwise, each action in the entire sequence predicted by the M2A model is penalized by a negative reward of $-1$.

\begin{figure}[ht]
    \centering
    \includegraphics[width=0.70\columnwidth]{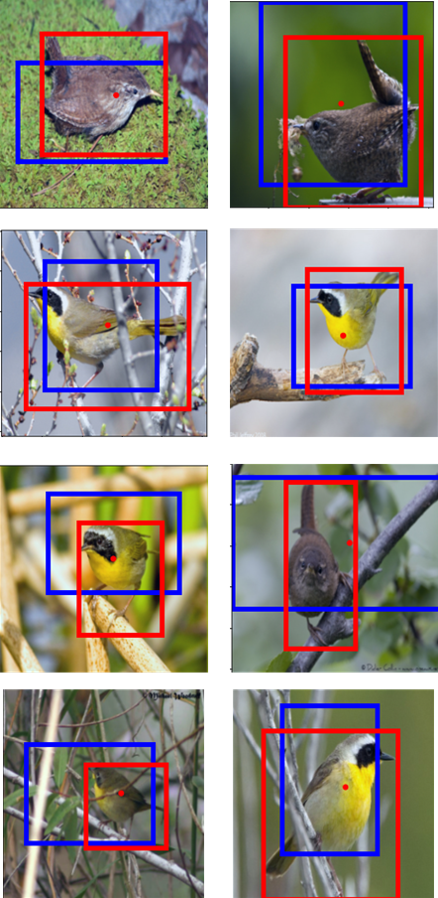}
    \vspace{-5pt}
    \caption{Randomly chosen examples from the test split of CUB dataset with the predicted bounding boxes (shown by blue rectangles), ground truth bounding boxes (shown by red rectangles) and initial fixation points (shown by red dots). Ground truth boxes are not used by the framework, but are only provided for reference purposes.\vspace{-6pt}}
    \label{fig:cub_testsamples}
\end{figure}

\subsubsection{\textbf{Quantitative and qualitative results}}
\input{Tables/Results_CUB_with_comparisons.tex}
\Cref{tab:birds_alt} and \Cref{fig:cub_testsamples} demonstrate the quantitative and the qualitative performance of our framework on the CUB~\cite{wah2011caltech} dataset, respectively.
If judged solely based on Top-1 Localization, the effectiveness of the proposed framework can be undervalued.
However, there are multiple factors that needs to be considered to evaluate the potential of the proposed method based on ``two-stream hypothesis" concepts.
\paragraph{Resource-constrained global context}
The other WSOL methods operate by relying on and improving upon the initial predictions made by the bounding box proposal methods, e.g. CAM or DDT.
Since such initial proposal methods have to still locate the entirety of the object of interest by predicting the bounding box dimensions, they have to process inputs in high details using large DNNs.
In contrast, our approach attempts to align with biological visual processes, within which the acquisition of the global context is fast and compute-efficient.
Hence, M1 model predicts the initial fixation points (requiring less details about object structures) instead of the initial bounding boxes (requiring more details about object structures). 
As a result, even though M1 model is realized using a primitive DNN model (refer to \cref{sec:res_cub_nnarchs}), it is capable of efficiently predicting approximate locations of objects of interest.
If we treat the centers of bounding boxes predicted by DDT as the initial fixation points, then the values of hit/miss accuracy shown in \cref{tab:birds_alt} support that the capability of M1 model to process the global context is matching that of DDT method.
\paragraph{Capturing the object of interest}
The qualitative results (shown in \cref{fig:cub_testsamples},~\cref{fig:cub_test_samples_extra_good}, and~\cref{fig:cub_test_samples_extra_bad}) reveal that, even when there is a misalignment between predicted and ground truth bounding boxes, the proposed framework is consistent at capturing the objects of interest.
This is also supported by the top-1 classification accuracy shown in \cref{tab:birds_alt}: since our proposed framework performs the prediction of the label \textit{based on the features extracted from the final foveated glimpse}, the results reveal that for $73.35\%$ of samples the final foveated glimpse captures the essential parts of the object required for its identification.
\paragraph{Probable cause of the limitation}
The comparison of the quantitative results (GT localization vs. hit/miss accuracy and top-1 classification accuracy) as well as the performance on CelebA dataset together suggest that the proposed framework falls short due to class labels. 
In particular, in the case of CelebA attributes, the framework operates by attempting to determine all of the object attributes, some of which require the entirety of the object to be captures.
Contrary, in the case of class labels of CUB dataset, the correct class confidence reaches a high value even when the object is observed partially, if that partial observation includes the most distinctive feature.
As a result, when the framework relies on class labels, and as the current foveated glimpse captures distinctive features, the M2A model loses ``incentive" (in the form of rewards) to explore more or to stop.
This is known as a ``part dominance" issue of weakly-supervised object localization task.
In the current formulation, this highlights the limitation of the proposed framework and the need for attributes as the guiding signal for the M2A model.
\paragraph{Novelty of the approach}
We want to emphasize the fact that the proposed approach is the first framework that learns independent localization function using the combination of supervised DNN training and reinforcement learning and explores the feasibility of the modeled mechanisms for machine perception. 
Existing WSOL works rely on the method of class activation map (CAM) and focus on improving it to capture entirety of the object as accurately as possible. 
Contrary, the neuroscience evidence suggests that the vision is an active process. 
Hence, our method illustrates the potential of using reinforcement learning in realizing the localization through the sequence of actions.
Furthermore, by deviating from the strictly neuro-inspired design, the methods of the proposed approach open opportunities to explore alternative hybrid strategies. For example, the work in~\cite{ibrayev2023exploring} investigates the integration of the modeled mechanisms of foveation and saccades with an existing WSOL method, showcasing their potential to enhance localization performance while tackling the issue of ``part dominance".


\subsection{Generalizability of dorsal stream model}
\label{sec:res_generalizability}
\begin{figure*}
    \centering
    \includegraphics[width=0.98\textwidth]{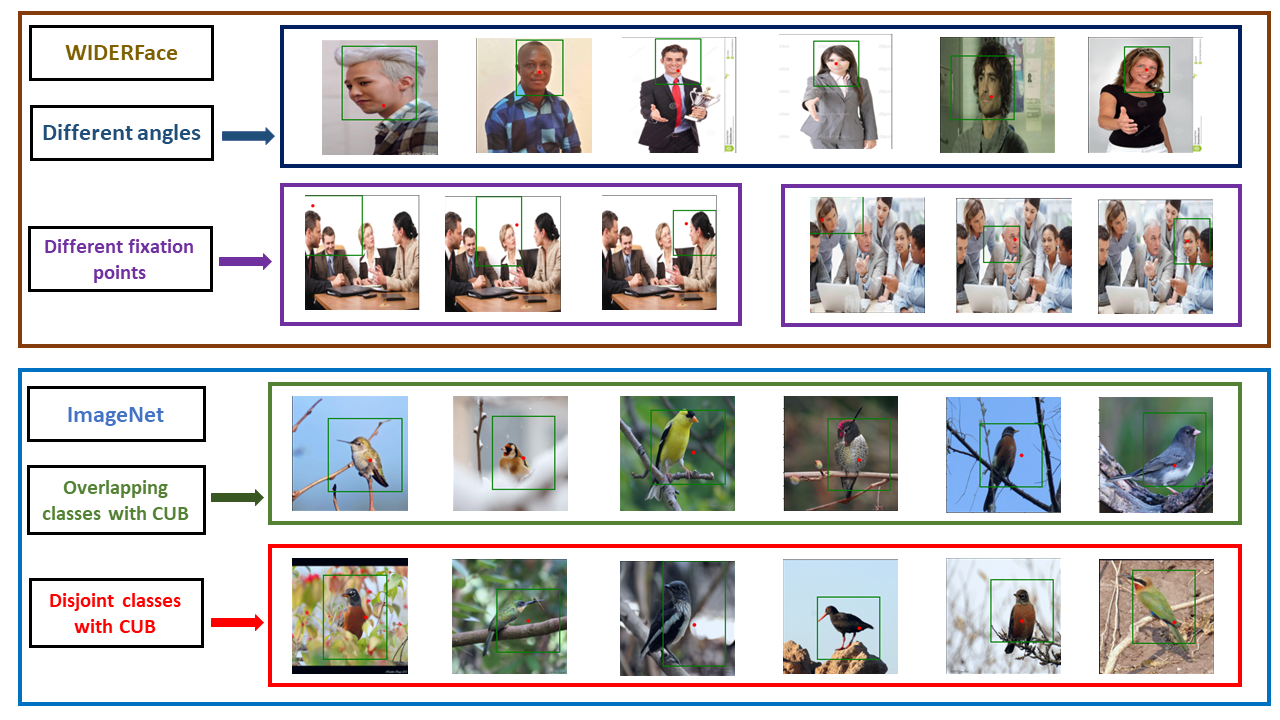}
    \vspace{-5pt}
    \caption{The generalization capabilities of the dorsal model. The top part shows the bounding boxes predicted (shown by green rectangles) on the unseen WIDERFace dataset predicted by the dorsal model trained on the CelebA faces dataset. The bottom part demonstrates similar generalization capabilities of the dorsal stream trained on CUB on a subset of bird classes sampled from the ImageNet dataset.\vspace{-3pt}}
    \label{fig:generalization}
\end{figure*}

One of the key advantages of the framework modeled based on ``\textit{two-stream hypothesis}" is the independence of each stream in terms of their performed functions.
Hence, here we illustrate the capability of the dorsal (where) stream model in terms of its generalization capabilities on completely unseen datasets. 
\Cref{fig:generalization} illustrates the qualitative results of the predicted localization on the WIDERFace \cite{yang2016wider} dataset and a subset of the ImageNet \cite{Deng2009ImageNetAL} dataset. A key point to emphasize here is that for inference, only the dorsal (where) model is being used with zero fine-tuning on these unseen datasets. The ventral (what) model is not used during these experiments.  

If we look into the top part of \cref{fig:generalization}, we can observe that the dorsal model detects near optimal bounding boxes (shown by green rectangles) on faces from challenging angles and is also able to detect different faces with different initial fixation points. It is really fascinating to note that for a different initial fixation point (shown as red dots) the dorsal model indeed comes up with a different localization result and this is a key difference between our and other localization approaches. In the bottom part of \cref{fig:generalization}, we observe that the dorsal model makes reasonable bounding box predictions (shown by green rectangles) on a subset of bird classes sampled from the ImageNet dataset. In the first row, we infer the predictions of partially overlapping classes between the CUB and the chosen subset from the ImageNet dataset and it shows the transferability properties of the dorsal stream. We make an even more exciting observation on the second row, where, for totally disjoint classes, the dorsal model localizes the object correctly even in the presence of extremely variable backgrounds.

%% file: Tables/Hyperparams_CelebA.tex
\renewcommand{\arraystretch}{0.87}
\begin{table*}
    \caption{Hyperparameters used to train \\ different components of the proposed framework on CelebA face attributes dataset}
    \label{tab:celeba_hyperparams}
    \centering
    \begin{tabular}{lccc}
        \toprule
         & \multicolumn{3}{c}{Training phase} \\
        \cmidrule(r){2-4} 
        Hyperparameter  & 
        \begin{tabular}[c]{@{}c@{}}Phase 0:\\pre-trained model\\(later used as M2B)\end{tabular} & 
        \begin{tabular}[c]{@{}c@{}}Phase 1:\\M1 model\\~~\end{tabular} & 
        \begin{tabular}[c]{@{}c@{}}Phase 2\&3:\\M2A model\\(w/ M2B model frozen)\end{tabular} \\
        \midrule
        Training epochs & 200 & 100 & 100 \\
        Training batch size & 128 & 128 & 128 \\
        Validation batch size & 50 & 100 & 100 \\
        Learning rate start & 1e-3 & 1e-2 & 1e-3 \\
        Learning rate min & 1e-5 & 1e-4 & 1e-5 \\
        \begin{tabular}[c]{@{}l@{}}Learning rate\\schedule\end{tabular} & 
        \begin{tabular}[c]{@{}c@{}}epochs\\\{100, 150\}\end{tabular} & 
        \begin{tabular}[c]{@{}c@{}}epochs\\\{50, 75\}\end{tabular} & 
        \begin{tabular}[c]{@{}c@{}}epochs\\\{50, 80\}\end{tabular}\\
        \begin{tabular}[c]{@{}l@{}}Learning rate\\drop factor\end{tabular} & 0.1 & 0.1 & 0.1 \\
        Weight decay & 0.0001 & 0.0 & 0.0001 \\
        \begin{tabular}[c]{@{}l@{}}Attribute detection\\threshold\end{tabular} & 0.5 & - & 0.5\\
        \midrule
        \begin{tabular}[c]{@{}l@{}}Number of glimpses\\($num\_glimpses$)\end{tabular} & - & - & 16\\
        \begin{tabular}[c]{@{}l@{}}Number of fovea neurons\\($fovea\_control\_neurons$)\end{tabular} & - & - & 5\\
        \begin{tabular}[c]{@{}l@{}}Initial glimpse size\\($glimpse\_size\_init$)\end{tabular} & - & - & 
        \begin{tabular}[c]{@{}c@{}}$20\times20$\\pixels\end{tabular}\\
        \begin{tabular}[c]{@{}l@{}}Resized glimpse size\\($glimpse\_size\_fixed$)\end{tabular} & - & - & \begin{tabular}[c]{@{}c@{}}$96\times96$\\pixels\end{tabular}\\
        \begin{tabular}[c]{@{}l@{}}Glimpse adjustment step\\($glimpse\_size\_step$)\end{tabular} & - & - & \begin{tabular}[c]{@{}c@{}}$20\times20$\\pixels\end{tabular}\\
        \bottomrule
    \end{tabular}
\end{table*}
\renewcommand{\arraystretch}{1}

%% file: Tables/Results_CelebA.tex
\renewcommand{\arraystretch}{1.0}
\begin{table}
    \caption[]{Performance of the framework components on the test set of the CelebA face attributes dataset.}
    \label{tab:celeba}
    \centering
\begin{tabular}{|cl|ccc|}
\hline
\multicolumn{2}{|c|}{\begin{tabular}[c]{@{}c@{}}Framework\\ Component\end{tabular}} &
  \begin{tabular}[c]{@{}c@{}}Pre-trained\\ model\end{tabular} &
  \begin{tabular}[c]{@{}c@{}}M1 \\ model\end{tabular} &
  \begin{tabular}[c]{@{}c@{}}Entire \\ framework\end{tabular} \\ \hline \hline
\multicolumn{2}{|c|}{\begin{tabular}[c]{@{}c@{}}Attribute \\ Accuracy\end{tabular}}     & 89.38 & -     & \begin{tabular}[c]{@{}c@{}}87.83 \\ $\pm$ 0.03\end{tabular} \\ \hline
\multicolumn{2}{|c|}{\begin{tabular}[c]{@{}c@{}}Hit/Miss\\ Accuracy\end{tabular}}   & -     & 88.70 & 88.70                                                   \\ \hline
\multicolumn{2}{|c|}{\begin{tabular}[c]{@{}c@{}}GT \\ Localization\end{tabular}}        & -     & -     & \begin{tabular}[c]{@{}c@{}}72.19 \\ $\pm$ 1.19\end{tabular} \\ \hline
\multicolumn{2}{|c|}{\begin{tabular}[c]{@{}c@{}}Attribute \\ Localization\end{tabular}} & -     & -     & \begin{tabular}[c]{@{}c@{}}63.63 \\ $\pm$ 1.05\end{tabular}  \\ \hline
\end{tabular}
\vspace{-9pt}
\end{table}
\renewcommand{\arraystretch}{1}

%% file: Tables/Hyperparams_CUB.tex
\renewcommand{\arraystretch}{0.87}
\begin{table*}
    \caption{Hyperparameters used to train \\different components of the proposed framework on CUB-200-2011 bird species dataset}
    \label{tab:cub_hyperparams}
    \centering
    \begin{tabular}{lccc}
        \toprule
         & \multicolumn{3}{c}{Training phase} \\
        \cmidrule(r){2-4} 
        Hyperparameter  & 
        \begin{tabular}[c]{@{}c@{}}Phase 0:\\pre-trained model\\(later used as M2B)\end{tabular} & 
        \begin{tabular}[c]{@{}c@{}}Phase 1:\\M1 model\\~~\end{tabular} & 
        \begin{tabular}[c]{@{}c@{}}Phase 2\&3:\\M2A model\\(w/ M2B model frozen)\end{tabular} \\
        \midrule
        Training epochs & 100 & 100 & 100 \\
        Training batch size & 64 & 128 & 32 \\
        Validation batch size & 50 & 100 & 50 \\
        Learning rate start & 1e-1 & 1e-2 & 1e-2 \\
        Learning rate min & 1e-5 & 1e-4 & 1e-5 \\
        \begin{tabular}[c]{@{}l@{}}Learning rate\\schedule\end{tabular} & 
        \begin{tabular}[c]{@{}c@{}}epochs\\\{30, 60, 90\}\end{tabular} & 
        \begin{tabular}[c]{@{}c@{}}epochs\\\{50, 75\}\end{tabular} & 
        \begin{tabular}[c]{@{}c@{}}epochs\\\{30, 60, 90\}\end{tabular}\\
        \begin{tabular}[c]{@{}l@{}}Learning rate\\drop factor\end{tabular} & 0.1 & 0.1 & 0.1 \\
        Weight decay & 0.0005 & 0.0 & 0.0001 \\
        \begin{tabular}[c]{@{}l@{}}Attribute detection\\threshold\end{tabular} & 0.5 & - & 0.5\\
        \midrule
        \begin{tabular}[c]{@{}l@{}}Number of glimpses\\($num\_glimpses$)\end{tabular} & - & - & 16\\
        \begin{tabular}[c]{@{}l@{}}Number of fovea neurons\\($fovea\_control\_neurons$)\end{tabular} & - & - & 5\\
        \begin{tabular}[c]{@{}l@{}}Initial glimpse size\\($glimpse\_size\_init$)\end{tabular} & - & - & 
        \begin{tabular}[c]{@{}c@{}}$80\times80$\\pixels\end{tabular}\\
        \begin{tabular}[c]{@{}l@{}}Resized glimpse size\\($glimpse\_size\_fixed$)\end{tabular} & - & - & \begin{tabular}[c]{@{}c@{}}$160\times160$\\pixels\end{tabular}\\
        \begin{tabular}[c]{@{}l@{}}Glimpse adjustment step\\($glimpse\_size\_step$)\end{tabular} & - & - & \begin{tabular}[c]{@{}c@{}}$20\times20$\\pixels\end{tabular}\\
        \begin{tabular}[c]{@{}l@{}}Similarity change threshold\\($sim\_change\_th$)\end{tabular} & - & - & 0.1\\
        \begin{tabular}[c]{@{}l@{}}Minimum satisfactory similarity\\($sim\_min\_satisfactory$)\end{tabular} & - & - & 0.5\\
        \bottomrule
    \end{tabular}
\end{table*}
\renewcommand{\arraystretch}{1}

%% file: Tables/Results_CUB_with_comparisons.tex
\renewcommand{\arraystretch}{1.2}
\begin{table*}[t]
\caption[]{Performance of the proposed framework and the other WSOL methods on the test set of the CUB-200-2011 birds dataset.}
\label{tab:birds_alt}
\centering
\begin{tabular}{|l|l|c|cc|c|}
\hline
\multicolumn{1}{|c|}{\multirow{3}{*}{Method}} & \multicolumn{1}{c|}{\multirow{3}{*}{Backbone network}} & \multirow{3}{*}{\begin{tabular}[c]{@{}c@{}}Top-1 Class\\ Accuracy\end{tabular}} & \multicolumn{2}{c|}{Localization Only} & \multirow{3}{*}{\begin{tabular}[c]{@{}c@{}}Top-1\\ Localization\end{tabular}} \\ \cline{4-5}
\multicolumn{1}{|c|}{} & \multicolumn{1}{c|}{} &  & \multicolumn{1}{c|}{\begin{tabular}[c]{@{}c@{}}Hit/Miss\\ Accuracy*\end{tabular}} & \begin{tabular}[c]{@{}c@{}}GT\\ Localization\end{tabular} &  \\ \hline
CAM~\cite{zhou2016learning} & VGG16 & - & \multicolumn{1}{c|}{-} & 57.96 & 36.13 \\ \hline
ADL~\cite{choe2019attention} & ResNet50 & 75.0 & \multicolumn{1}{c|}{-} & 77.60 & 59.50 \\ \hline
SLT-Net~\cite{comptable_SLTNET} & InceptionV3 & 76.4 & \multicolumn{1}{c|}{-} & 86.50 & 66.10 \\ \hline
DDT~\cite{wei2019unsupervised} & VGG16 & NA & \multicolumn{1}{c|}{99.64} & 88.51 & 62.30 \\ \hline
\multirow{2}{*}{\begin{tabular}[c]{@{}l@{}}PSOL~\cite{comptable_PSOL}\\ (w/  DDT as initial bounding box predictor)\end{tabular}} & \multirow{2}{*}{\begin{tabular}[c]{@{}l@{}}VGG16 (localization) +\\ ResNet50 (classification)\end{tabular}} & \multirow{2}{*}{75.0} & \multicolumn{1}{c|}{\multirow{2}{*}{99.64}} & \multirow{2}{*}{77.41} & \multirow{2}{*}{63.56} \\
 &  &  & \multicolumn{1}{c|}{} &  &  \\ \hline
\multirow{2}{*}{\begin{tabular}[c]{@{}l@{}}Proposed framework\\ (w/ M1 as initial fixation predictor)\end{tabular}} & \multirow{2}{*}{\begin{tabular}[c]{@{}l@{}}ResNet50 (M2A) + \\ ResNet50 (M2B)\end{tabular}} & \multirow{2}{*}{73.35} & \multicolumn{1}{c|}{\multirow{2}{*}{93.33}} & \multirow{2}{*}{38.32} & \multirow{2}{*}{29.18} \\
 &  &  & \multicolumn{1}{c|}{} &  &  \\ \hline
\end{tabular}
\vspace{1mm}
\\ \footnotesize{$^*$ \textit{\textbf{Note:} ``Hit/Miss Accuracy" is estimated only for DDT method and M1 model. For DDT, it is the ratio of samples for which the center of the bounding box predicted by DDT is within the GT bounding box. For M1 model, it is the ratio of samples for which the initial fixation point is within the GT bounding box.}}
\vspace{-4mm}
\end{table*}
\renewcommand{\arraystretch}{1.0}

%% file: Sections/5-Conclusion.tex
\section{Conclusion \& Future Directions}
While the current computer vision frameworks achieved significant performance on complex datasets, there are still missing gaps between machine perception and human vision.
In this work, we proposed a machine learning framework inspired by the ``\textit{two-stream hypothesis}" with the goal of exploring its potential in guiding the development of bio-plausible methods for machine perception.
We presented a deep neural network-based framework, modeling the two stream hypothesis (what and where of vision).
The ventral (what) stream is modeled as a neural network trained, in a supervised manner to predict the object and its parts making use of the foveation mechanism.
The dorsal (where) stream is modeled as a separate neural network trained via reinforcement learning to predict the context and location of the object.
The framework then combines the two stream models into iterative learning.

The feasibility of the framework as the model for visual recognition and localization was illustrated on the task of weakly-supervised object localization on the CelebA and CUB datasets.
The results indicate that the foveation forces the framework to separate the object parts from the background clutter.
The combination of the two streams enables the framework to learn the correct sequence of focused observations, leading it to successfully locate and capture the objects of interest. 
The results also highlight the generalization properties of the dorsal stream.
Specifically, the dorsal stream model is able to localize objects on completely unseen data without any need for additional re-training.
We believe that incorporating bio-plausible mechanisms like foveation into the future development of machine perception will pave the pathway for more interesting learning mechanism, closing the gap between cognitive and machine intelligence.

%% file: Sections/Appendix.tex
\section{Additional qualitative samples from experiments on CelebA dataset}
\label{sec:appendix_celeba}

\begin{figure}
    \centering
    \includegraphics[width=0.8\columnwidth]{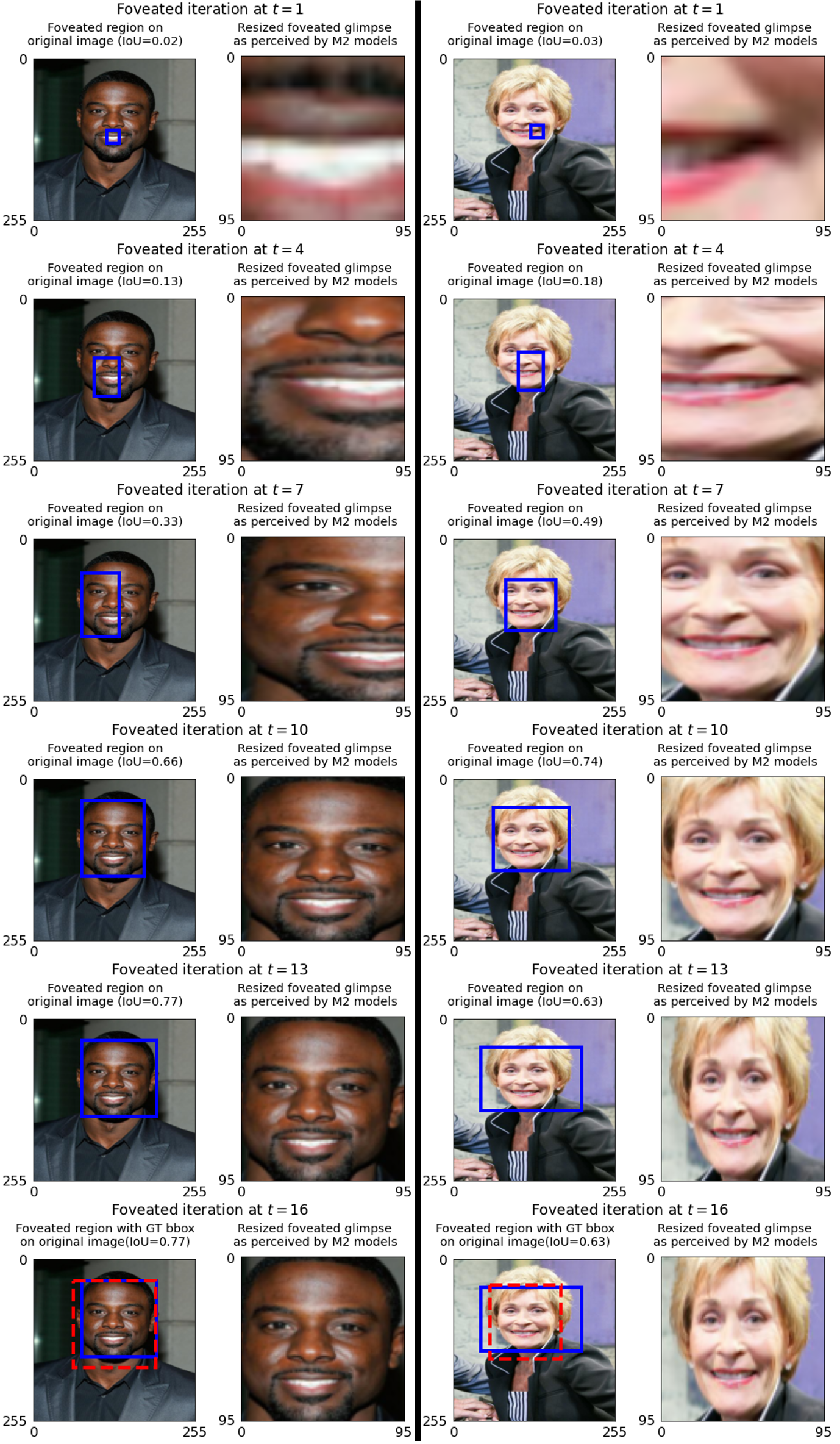}
    \caption{Extra test samples illustrating successful localization, i.e. $IoU \geq 0.5$ with GT bbox (shown as red dashed rectangles, only for reference purposes)}
    \label{fig:celeba_test_samples_extra_good}
\end{figure}
\begin{figure}
    \centering
    \includegraphics[width=0.8\columnwidth]{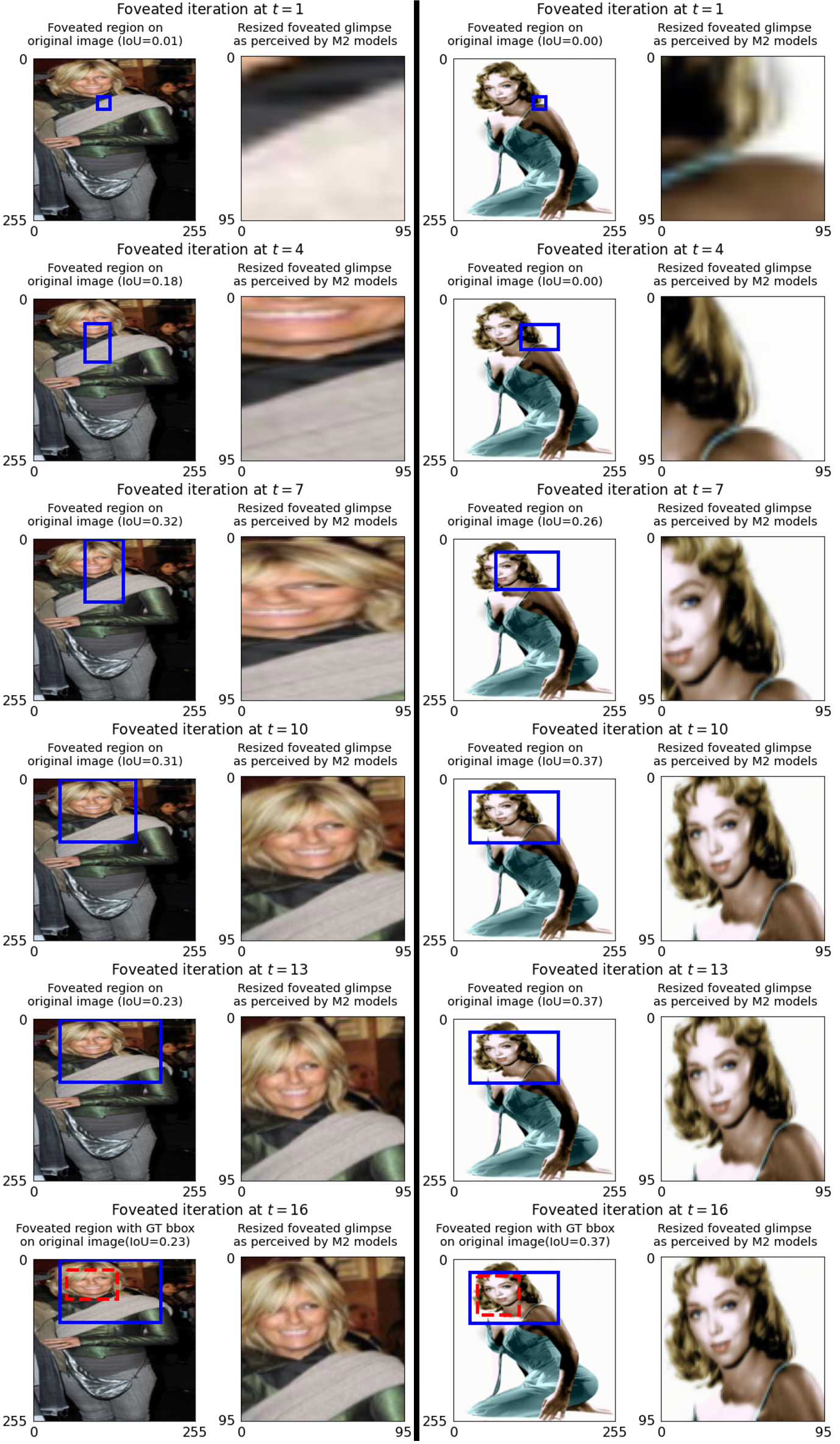}
    \caption{Extra test samples illustrating failed localization, i.e. $IoU < 0.5$ with GT bbox (shown as red dashed rectangles, only for reference purposes)}
    \label{fig:celeba_test_samples_extra_bad}
\end{figure}

\Cref{fig:celeba_test_samples_extra_good,fig:celeba_test_samples_extra_bad} illustrate extra test samples that have intersection over union with ground truth bounding box (GT bbox) above $0.5$ and below $0.5$, respectively.
Example cases presented in \cref{fig:celeba_test_samples_extra_bad} show that one of the situations where localization fails is due to non-optimal prediction of initial fixation point. 
Despite the framework still able to localize the face, because of lack of shrink actions, the final glimpse captures face region with excessive background clutter.
More of the samples can be found within supplementary material (which will later become available as a part of code repository).

\section{Additional qualitative samples from experiments on CUB-200-2011 dataset}
\begin{figure}
    \centering
    \includegraphics[width=0.80\columnwidth]{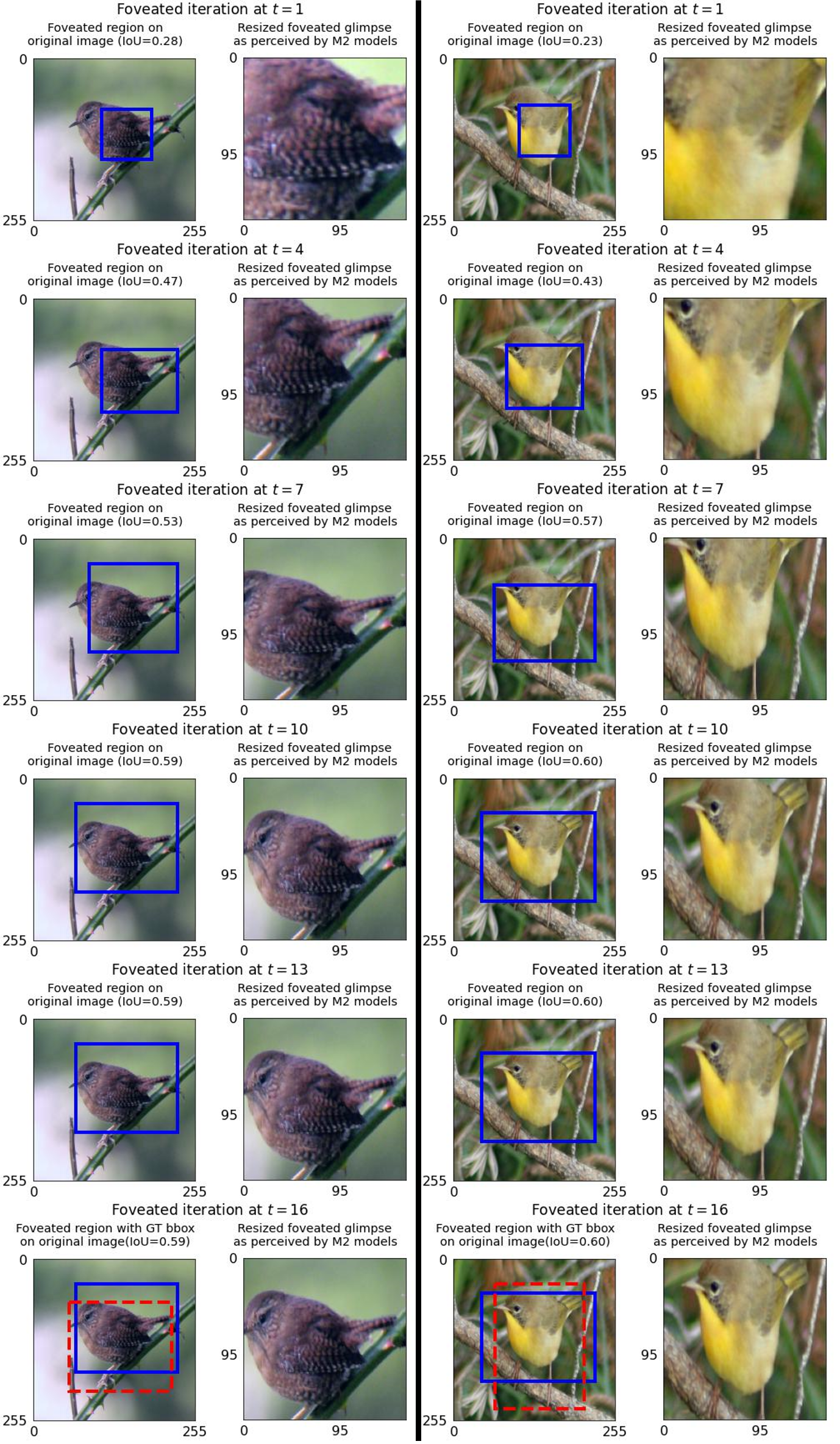}
    \caption{Extra test samples illustrating successful localization, i.e. $IoU \geq 0.5$ with GT bbox (shown as red dashed rectangles, only for reference purposes)}
    \label{fig:cub_test_samples_extra_good}
\end{figure}

\begin{figure}
    \centering
    \includegraphics[width=0.80\columnwidth]{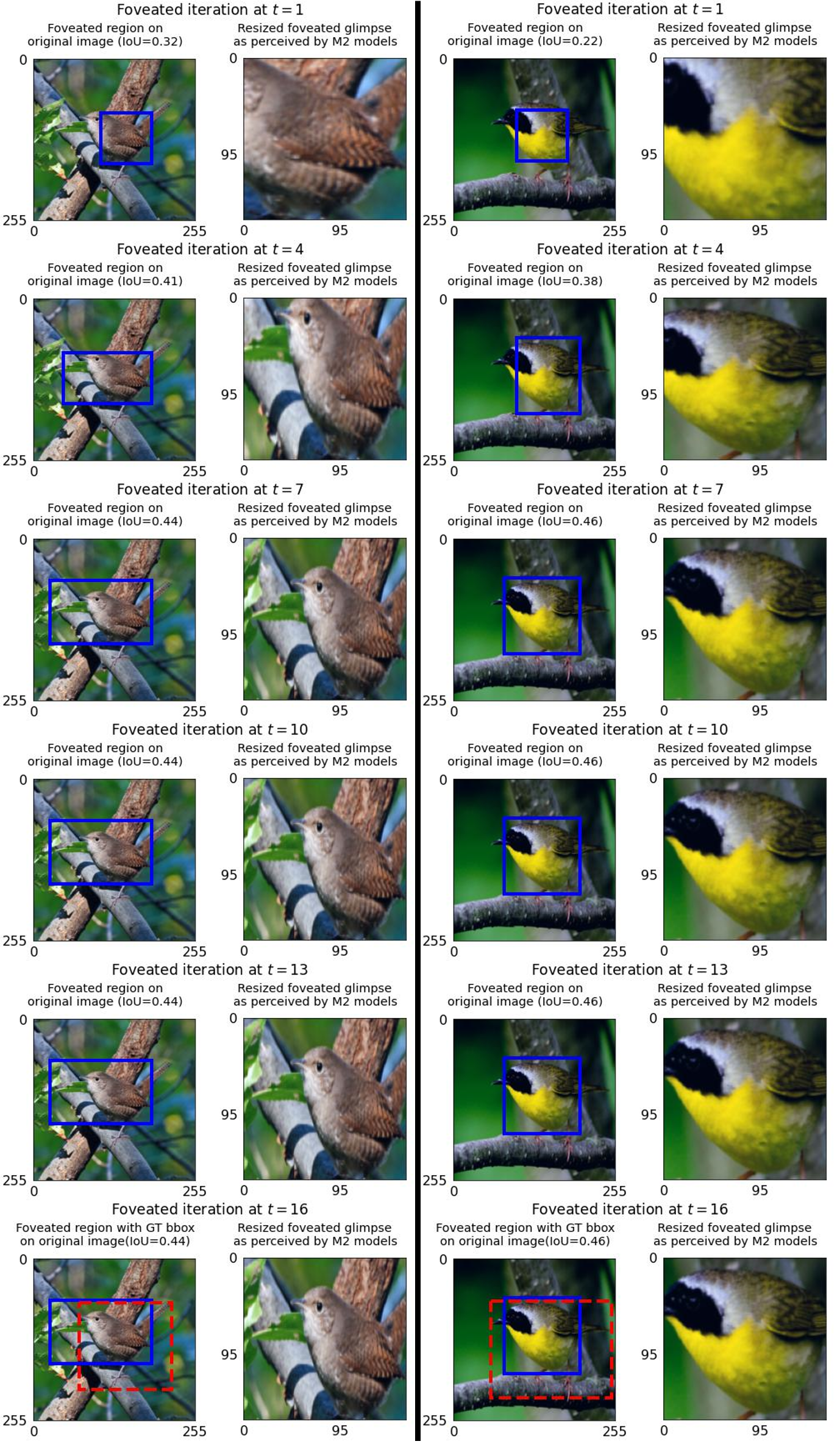}
    \caption{Extra test samples illustrating failed localization, i.e. $IoU < 0.5$ with GT bbox (shown as red dashed rectangles, only for reference purposes)}
    \label{fig:cub_test_samples_extra_bad}
\end{figure}

\Cref{fig:cub_test_samples_extra_good,fig:cub_test_samples_extra_bad} illustrate extra test samples that have intersection over union with ground truth bounding box (GT bbox) above $0.5$ and below $0.5$, respectively.
More of the samples can be found within supplementary material (which will later become available as a part of the code repository).

\section{Additional qualitative samples from generalization experiments}

\begin{figure*}
    \centering
    \includegraphics[width=1.70\columnwidth]{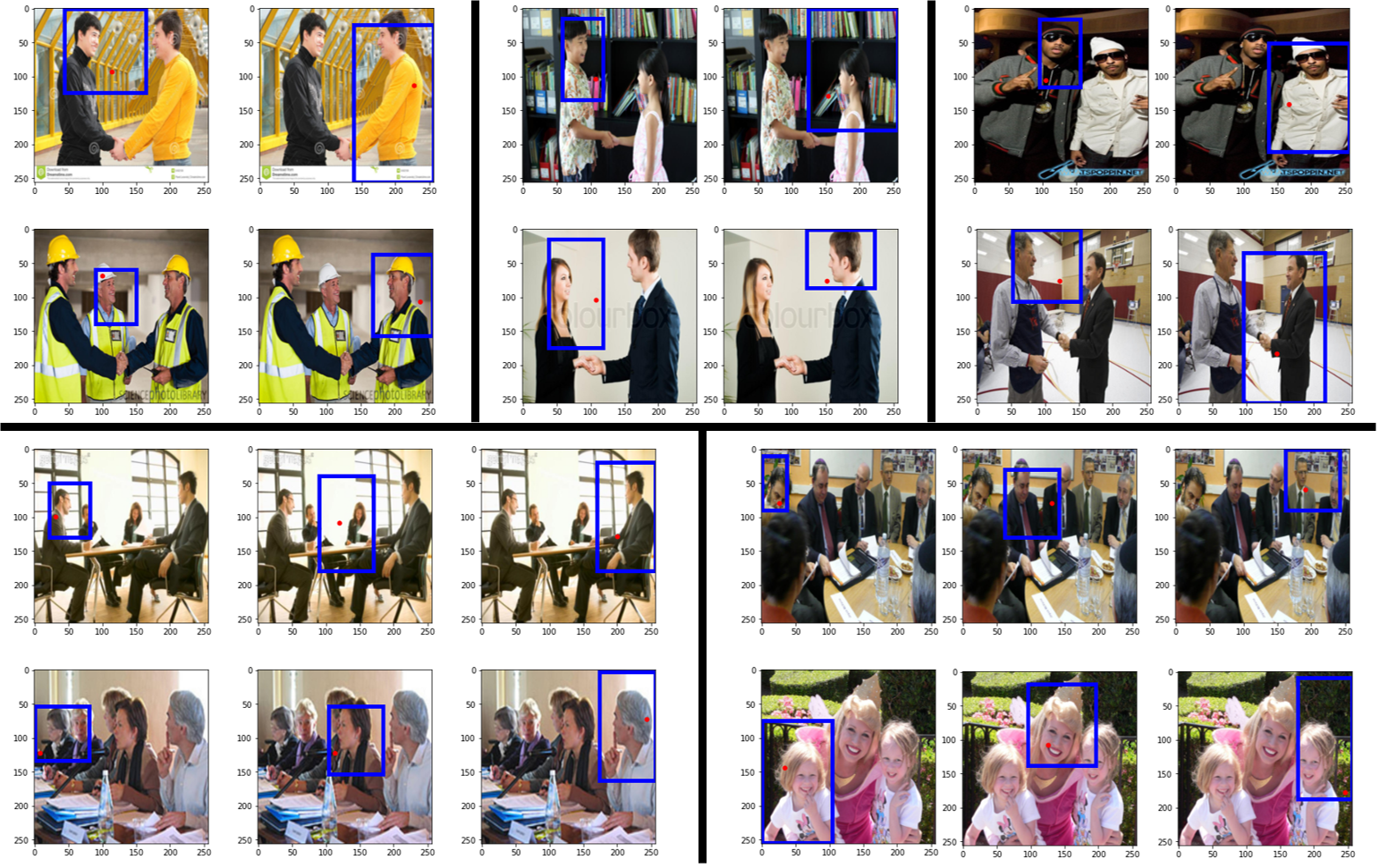}
    \caption{Extra test samples illustrating successful generalization of the dorsal stream model on unseen images from a different dataset (human faces from WIDERFace) when trained on CelebA dataset.}
    \label{fig:widerfaces_extra}
\end{figure*}

\Cref{fig:widerfaces_extra} illustrates extra results on localization of human faces from the WIDERFace dataset using only the dorsal stream model (i.e. M2A model) trained on the CelebA dataset.

\begin{figure}
    \centering
    \includegraphics[width=0.80\columnwidth]{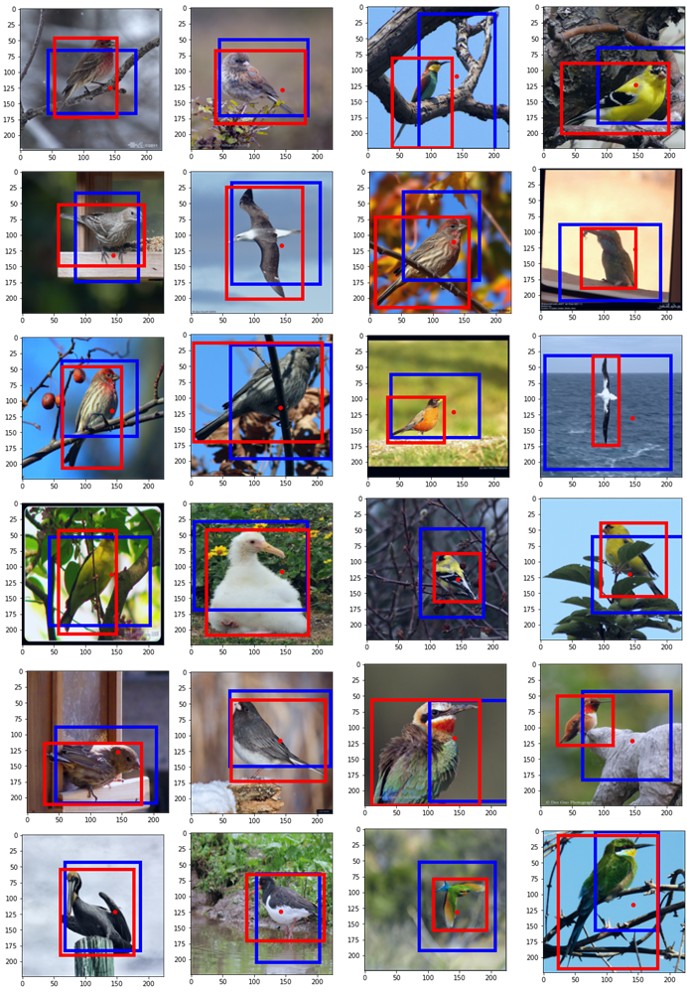}
    \caption{Extra test samples illustrating successful generalization of the dorsal stream model on unseen images from a different dataset (birds subset from ImageNet) when trained on CUB dataset. Red rectangles show the ground truth bounding boxes provided by the dataset, but which are not used by the framework.}
    \label{fig:imagenet_extra}
\end{figure}

\Cref{fig:imagenet_extra} illustrates extra results on localization of birds subset from ImageNet dataset using only dorsal stream model (i.e. M2A model) trained on CUB dataset.


%% file: Sections/AuthorBios.tex
\begin{IEEEbiography}[{\includegraphics[width=1in,height=1.25in,clip,keepaspectratio]{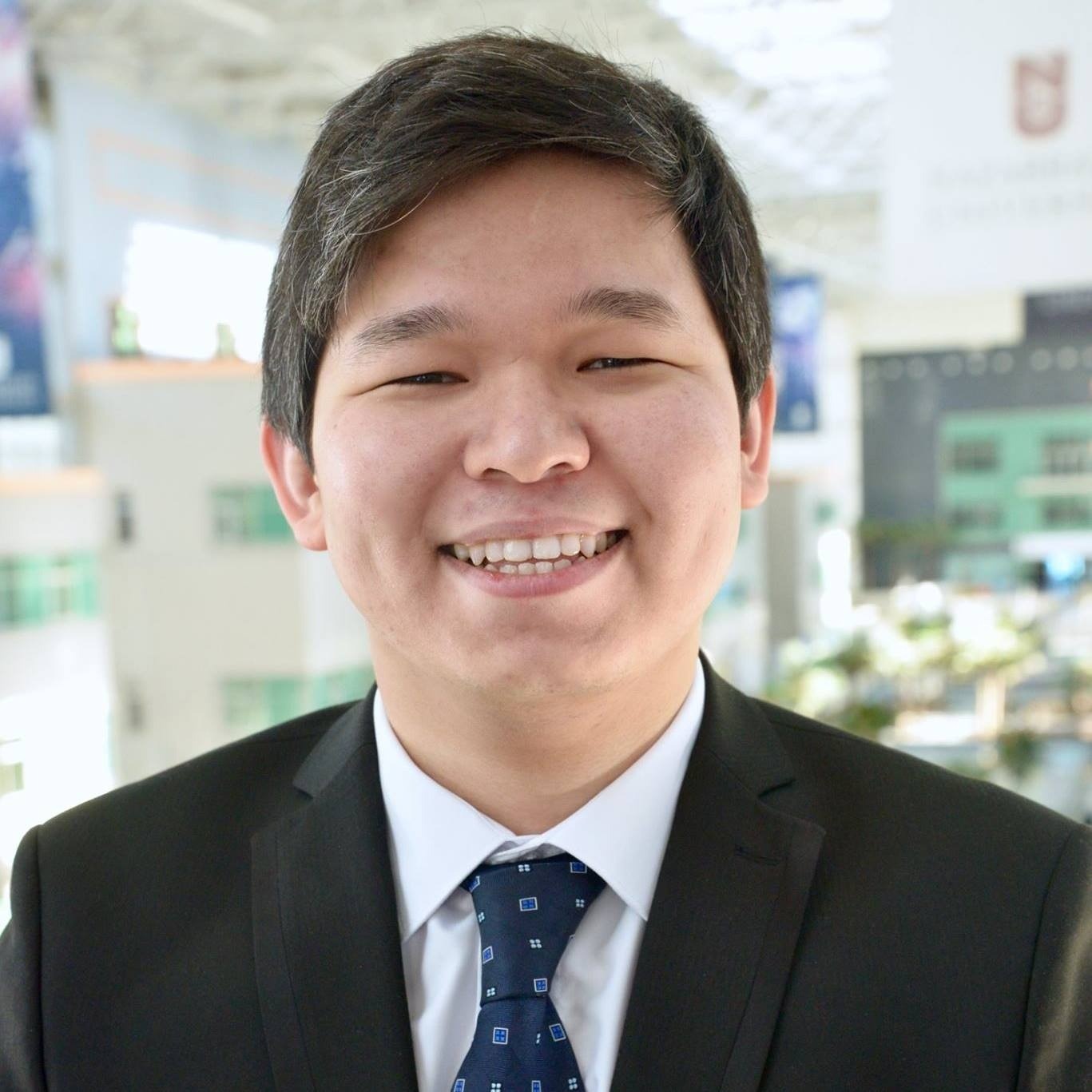}}]
{Timur Ibrayev} received the B.S. in electrical and electronics engineering from Nazarbayev University, Astana, Kazakhstan, in 2017. He is currently pursuing the Ph.D. degree in electrical and computer engineering at Purdue University, West Lafayette, IN, USA. He interned at Intel Corp. during Summer 2022, where he worked on profiling and optimization of the deep learning code base for their clusters based on Gaudi deep learning processors. His research interests include hardware-software co-design and neuro-inspired computer vision algorithms.
\end{IEEEbiography}

\begin{IEEEbiography}[{\includegraphics[width=1in,height=1.25in,clip,keepaspectratio]{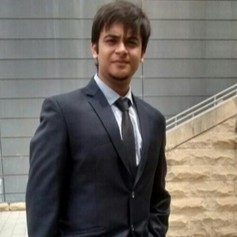}}]
{Amitangshu Mukherjee} earned his B.Tech in Applied Electronics and Instrumentation Engineering from West Bengal University of Technology in 2016. He completed his Master’s in Computer Engineering at Iowa State University, USA, in Fall 2019. His research primarily centered on exploring Deep Learning-based algorithms for problems in domain adaptation, generative learning, and adversarial deep learning, with a particular emphasis on their applications in robotic perception systems. In summer 2020, he commenced his PhD studies at Purdue University’s ECE department, supervised by Professor Kaushik Roy at the Center for Brain-Inspired Computing (C-BRIC). His doctoral research primarily revolves around understanding and designing neuro-inspired algorithms for robust learning in autonomous systems. During summer 2023, Amitangshu interned at Qualcomm Inc.'s Multimedia Research and Development (MMRND) team, where he contributed to the development of a novel multi-tasking transformer-based architecture for optical flow and depth estimation for autonomous driving scenes.
\end{IEEEbiography}

\begin{IEEEbiography}[{\includegraphics[width=1in,height=1.25in,clip,keepaspectratio]{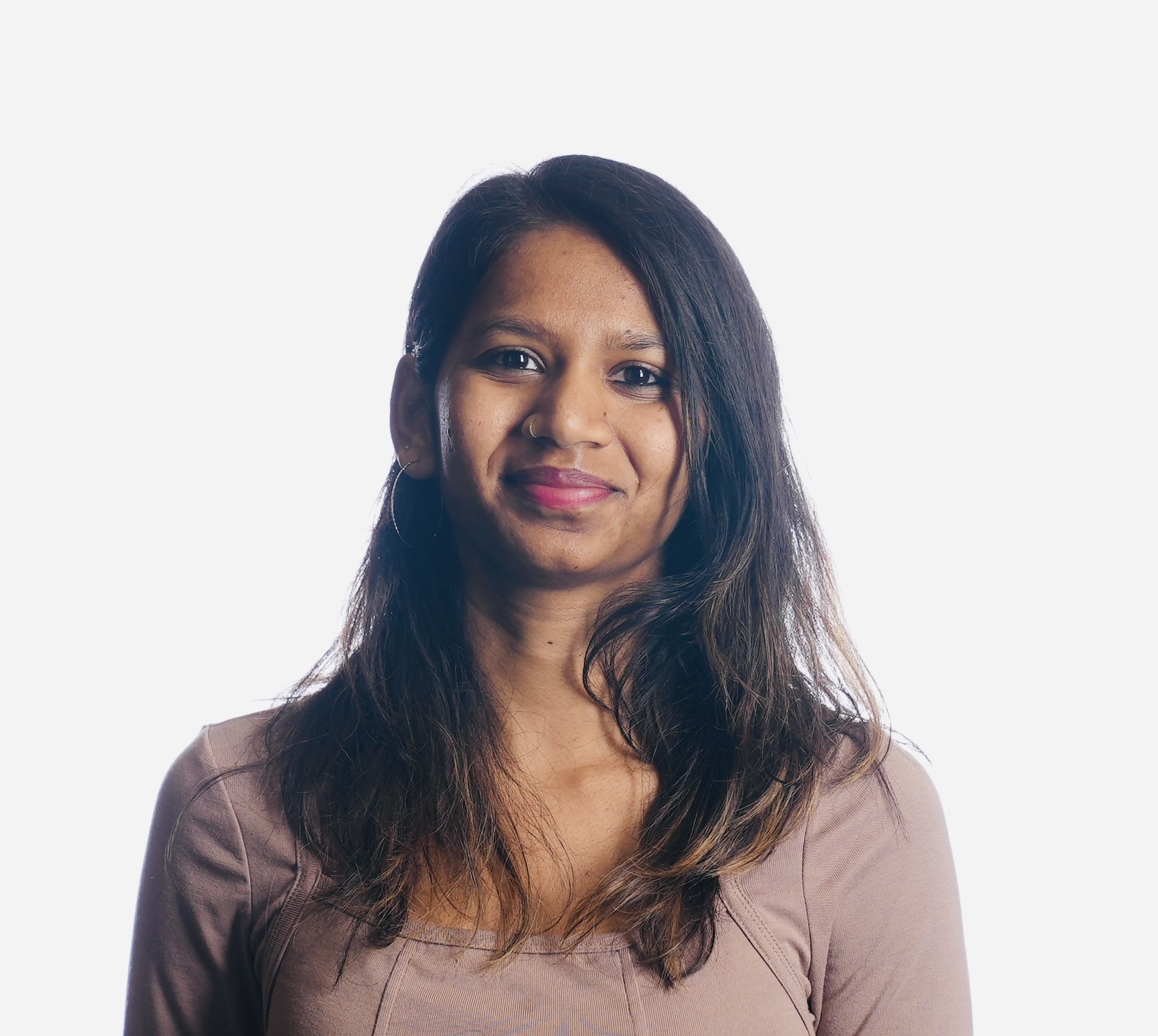}}]
{Aparna Aketi} received her B.Tech. in Electrical Engineering from the Indian Institute of Technology, Gandhinagar, India, in 2018. She joined Purdue in the Fall of 2018 as a Ph.D. student under the guidance of Professor Kaushik Roy. In the initial years of her Ph.D., she explored various topics related to efficient machine learning methods such as pruning, early exits, and low-precision training. Her research interests include privacy-preserving machine learning, federated learning, and decentralized optimization. Currently, she is working on developing algorithms for decentralized (peer-to-peer) learning setups that support heterogeneous data distributions.
\end{IEEEbiography}

\begin{IEEEbiography}[{\includegraphics[width=1in,height=1.25in,clip,keepaspectratio]{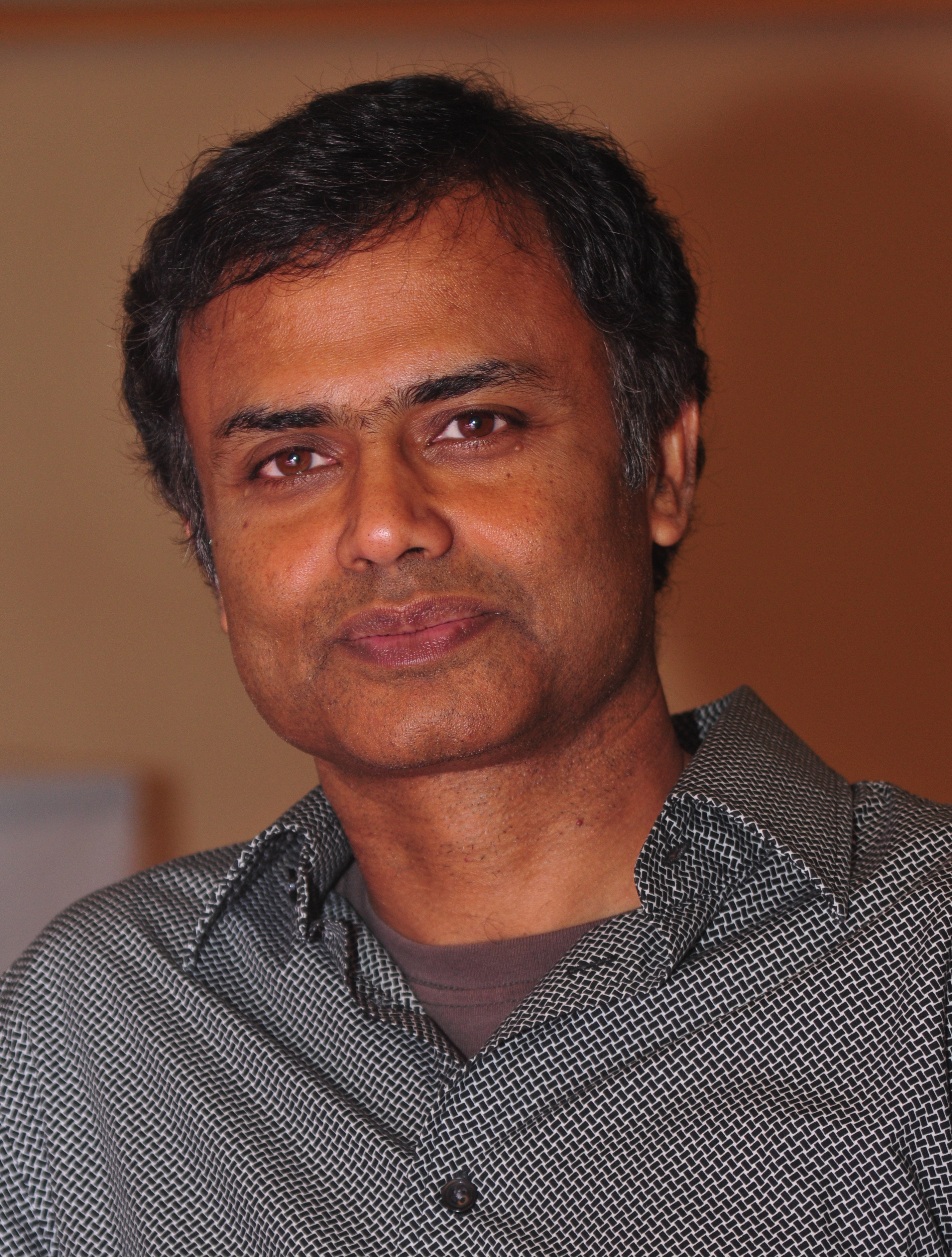}}]
{Kaushik Roy (Fellow, IEEE)}  received the B.Tech. degree in electronics and electrical communications engineering from IIT Kharagpur, Kharagpur, India, and the Ph.D. degree from the electrical and computer engineering department, University of Illinois at Urbana–Champaign, in 1990.

He was with the Semiconductor Process and Design Center of Texas Instruments, Dallas, where
he worked on FPGA architecture development and low-power circuit design. He joined the Electrical and Computer Engineering Faculty, Purdue University, West Lafayette, IN, USA, in 1993, where he is
currently the Edward G. Tiedemann Jr. Distinguished Professor. He is the Director of the Center for Brain-Inspired Computing (C-BRIC) funded by SRC/DARPA. He has published more than 700 articles in refereed journals and conferences, holds 25 patents, supervised more than 100 Ph.D. dissertations, and is co-author of two books on Low Power CMOS VLSI Design (John Wiley \& McGraw Hill). His research interests include neuromorphic and emerging computing models, neuro-mimetic devices, spintronics, device-circuitalgorithm co-design for nano-scale silicon, non-silicon technologies, and low-power electronics. He received the National Science Foundation Career Development Award, in 1995, the IBM Faculty Partnership Award, the ATT/Lucent Foundation Award, the 2005 SRC Technical Excellence Award,
the SRC Inventors Award, the Purdue College of Engineering Research Excellence Award, the Humboldt Research Award, in 2010, the 2010 IEEE Circuits and Systems Society Technical Achievement Award (Charles Desoer Award), the Distinguished Alumnus Award from IIT Kharagpur, the Fulbright-Nehru Distinguished Chair, the DoD Vannevar Bush Faculty Fellow, from 2014 to 2019, the Semiconductor Research Corporation Aristotle Award, in 2015, the 2020 Arden Bement Jr. Award, the Highest Research
Award given by Purdue University in pure and applied science and engineering, and the Best Paper Awards at 1997 International Test Conference, the IEEE 2000 International Symposium on Quality of IC Design, the 2003 IEEE Latin American Test Workshop, the 2003 IEEE Nano, the 2004 IEEE International Conference on Computer Design, the 2006 IEEE/ACM International Symposium on Low Power Electronics and Design, and the 2005 IEEE Circuits and System Society Outstanding Young Author Award (Chris Kim), the 2006 IEEE TRANSACTIONS ON VLSI SYSTEMS Best Paper Award, the 2012 ACM/IEEE International Symposium on Low Power Electronics and Design Best Paper Award, and the 2013 IEEE TRANSACTIONS ON VLSI Best Paper Award. He was a Purdue University Faculty Scholar from 1998 to 2003. He was a Research Visionary Board Member of Motorola Labs, in 2002 and held the M. Gandhi Distinguished Visiting Faculty with IIT Bombay and the Global Foundries Visiting Chair with the National University of Singapore. He has been in the editorial board of IEEE Design and Test, the IEEE TRANSACTIONS ON CIRCUITS AND SYSTEMS, the IEEE TRANSACTIONS ON
VLSI SYSTEMS, and the IEEE TRANSACTIONS ON ELECTRON DEVICES. He was a Guest Editor for Special Issue on Low-Power VLSI in IEEE Design and Test, in 1994, the IEEE TRANSACTIONS ON VLSI SYSTEMS, in June 2000, the IEEE Proceedings–Computers and Digital Techniques, in July 2002, and the IEEE JOURNAL ON EMERGING AND SELECTED TOPICS IN CIRCUITS AND SYSTEMS, in 2011. 
\end{IEEEbiography}